\documentclass[twoside]{article}

\usepackage[preprint]{aistats2027}
\usepackage[round]{natbib}

\usepackage{hyperref}       % hyperlinks
\usepackage{url}            % simple URL typesetting
\usepackage{booktabs}       % professional-quality tables
\usepackage{amsfonts}       % blackboard math symbols
\usepackage{nicefrac}       % compact symbols for 1/2, etc.
\usepackage{microtype}      % microtypography
\usepackage[table,xcdraw]{xcolor} % Load required options once.

\usepackage{multirow}
\usepackage{tikz}
\usepackage{graphicx}
\usepackage{mathtools}
\usepackage{amssymb}
\usepackage{amsthm}
\usepackage{thmtools}
\usepackage{floatrow}
\newfloatcommand{capbtabbox}{table}[][\FBwidth]

\usepackage{amsmath}
\usepackage{mathtools}
\usepackage{amsfonts} % various math stuff
\usepackage{wrapfig}  % Allow wrapping of text around figures
\usepackage{mdwlist}  % Make list items closer together: itemize*, enumerate*
\usepackage{sidecap}  % For putting caption beside figure
\usepackage{bbm}
\usepackage{bm,amsbsy} % for bold symbols
\usepackage[normalem]{ulem} % For \sout (strikethorugh)
\usepackage{subcaption}

\usepackage{placeins} % for \FloatBarrier

\usepackage{algorithm}
\usepackage{algpseudocode}
\usepackage{booktabs}       % professional-quality tables
\usepackage{amsfonts}       % blackboard math symbols
\usepackage{nicefrac}       % compact symbols for 1/2, etc.
\usepackage{microtype}      % microtypography
\usepackage{multirow}
\usepackage{array}
\usepackage{graphicx}
\usepackage{standalone} 
\usepackage{adjustbox}
 \usepackage{colortbl}
\newcommand\scalemath[2]{\scalebox{#1}{\mbox{\ensuremath{\displaystyle #2}}}}

\usepackage{etoc}

\definecolor{poolyellow}{RGB}{255,255,102}
\definecolor{poolpink}{RGB}{244,194,204}
\definecolor{poolblue}{RGB}{173,216,230}
\definecolor{headgray}{RGB}{190,190,190}

\usepackage{tabularray}
\usepackage{ulem, bm, kotex}

\usepackage{lipsum}
\usepackage[most]{tcolorbox}
\definecolor{DefinitionBarColor}{gray}{0.35}  % 왼쪽 막대 색상 (중간-어두운 회색)
\definecolor{DefinitionBackColor}{gray}{0.95} % 배경 색상 (매우 연한 회색)

\newtcolorbox{mydefinitionbox}[1][]{
    enhanced, % 고급 그리기 옵션 사용
    breakable, % 페이지 넘어갈 때 박스 자동 분할
    boxrule=0pt,    % 전체 테두리 두께 (0pt로 설정하여 보이지 않게 함)
    leftrule=3pt,   % 왼쪽 막대의 두께
    rightrule=0pt,  % 오른쪽 선 없음
    toprule=0pt,    % 위쪽 선 없음
    bottomrule=0pt, % 아래쪽 선 없음
    arc=0mm,        % 모서리를 둥글게 하지 않음 (직각)
    boxsep=0pt,     % 기본 여백을 0으로 설정하고 아래에서 세부 조정
    top=1pt,        % 내용과 위쪽 경계 사이 여백
    bottom=1pt,     % 내용과 아래쪽 경계 사이 여백
    left=2pt,       % 내용과 왼쪽 막대 사이 여백
    right=2pt,      % 내용과 오른쪽 경계 사이 여백
    before skip=\medskipamount,
    after skip=\medskipamount,
    #1 % 추가적인 tcolorbox 옵션을 넘겨줄 수 있도록 함 (예: [title=나의 정의])
}

\usepackage{amsthm}
\usepackage{xcolor}
\usepackage{hyperref}
\usepackage{amsmath,amsfonts,bm}

\newcommand{\Nrm}{\mathcal{N}}
\renewcommand{\eqref}[1]{Eq.~(\ref{eq:#1})}

\newcommand{\figref}[1]{Fig.~\ref{fig:#1}}  % use for citing figs

\newcommand{\tabref}[1]{Table ~\ref{tab:#1}}  % use for citing tables
\newcommand{\algoref}[1]{Algorithm~\ref{algo:#1}}  % use for citing algorithms

\newcommand{\suppsecref}[1]{Supplementary Sec.~\ref{supp:#1}}

\newcommand{\Dat}{\mathcal{D}}
\newcommand{\valpha}{\mathbf{\ensuremath{\bm{\alpha}}}}
\newcommand{\vbeta}{\mathbf{\ensuremath{\bm{\beta}}}}
\newcommand{\vphi}{\mathbf{\ensuremath{\bm{\phi}}}}

\newtheorem{lem}{Lemma}[section]

\def\1{\bm{1}}

\def\vtheta{{\bm{\theta}}}

\def\vw{{\bm{w}}}
\def\vx{{\bm{x}}}
\def\vy{{\bm{y}}}

\DeclareMathAlphabet{\mathsfit}{\encodingdefault}{\sfdefault}{m}{sl}
\SetMathAlphabet{\mathsfit}{bold}{\encodingdefault}{\sfdefault}{bx}{n}

\DeclareMathOperator{\diag}{diag} % Used in the supplementary prior covariance.
\newcommand{\tr}{{}^{\top}} % Restore the transpose shorthand used in the appendix.

\begin{document}



\twocolumn[

\aistatstitle{Optimizable Model Averaging\\
for Ensembling Heterogeneous Model Outputs}

\aistatsauthor{Mijung Park}

\aistatsaddress{mijungp@cs.ubc.ca, Computer Science, UBC Vancouver } 
]

%%%% three key points to convey in this paper
%%% (1) OMA: black-box model output averaging approach, based on predictive entropy, to adapt to unseen test data distribution
%%% (2) predictive entropy itself isn't enough. To encourage merged model outputs to be better calibrated, we suggest KL regularization for text classification  
%%% (3) additionally implementing Bayesian model averaging for image foundation models, which serves as informative-prior for OMA, regularizing the model ensembling weights   

\begin{abstract}
Foundation model ensembles can improve accuracy and robustness, but full fine-tuning and parameter-space merging are often expensive or impossible when candidate models have different architectures. 
We propose \textit{Optimizable Model Averaging} (OMA), a black-box approach that learns nonnegative model-combination weights, one scalar coefficient per model, to average the candidate models’ predictive probabilities, leaving all model parameters frozen, by minimizing regularized predictive entropy. 
A central challenge is that minimizing predictive entropy can favour confidently incorrect models: increased confidence need not imply improved accuracy or calibration. 
We therefore examine regularization as a mechanism for controlling adaptation and empirically evaluate its effect on calibration, observing lower expected calibration error in selected configurations, alongside better performance–calibration trade-offs than model output averaging.
As foundation models continue to advance, these approaches will enable the incorporation of future, potentially much better models to improve performance on challenging text and image classification tasks.
\end{abstract}

\section{Introduction}\label{sec:intro}

Foundation models provide strong starting points for image and text
classification, but adapting them to new data remains challenging.
In vision, models such as CLIP support zero-shot classification
through transferable image--text representations
\citep{radford2021learning}, while subsequent adaptation can further
improve downstream performance
\citep{yu2022coca,EVA02L,zanella2024low}. However, fine-tuning multiple
large models can require substantial computation and access to
their parameters. Ensembling offers an alternative way to exploit
existing models and their complementary predictions. In particular,
combining zero-shot and adapted models can help balance
task-specific performance and robustness under distribution shift
\citep{wortsman2021robust,andreassen2022the}.

Two common approaches are parameter-space merging and output
averaging. Parameter-space methods, such as Model Soups
\citep{model_soups}, combine compatible checkpoints into a single
model, avoiding the additional inference cost of output ensembling.
However, their applicability is limited, when candidate models have
different architectures or their parameters are inaccessible.
Output averaging instead combines predictive distributions and can
therefore accommodate heterogeneous models, provided that their
probabilities refer to a shared label space. Uniform averaging is a
useful baseline, but assigns the same contribution to every model
regardless of the target distribution. Source-derived weights can
account for differences on the source data, yet may become less
informative when model performance changes under distribution shift.
This motivates our central question:
\emph{How can we adapt the contributions of heterogeneous models
using only their predictions on unlabeled target data?}

We propose \emph{Optimizable Model Averaging} (OMA), a black-box
approach to target-adaptive output ensembling. Given $L$ candidate
predictors, with the $l$th model's predictive distribution set to $p_\ell(\vy\mid \vx)$, OMA forms the ensembled predictive distribution $q_{\vbeta}(\vy\mid \vx)$ with a vector of mixture weights $\vbeta$:
\begin{equation}
    q_{\boldsymbol{\beta}}(\vy\mid \vx)
    = \sum_{\ell=1}^{L} \beta_\ell p_\ell(\vy\mid \vx),
    \label{eq:intro-oma-mixture}
\end{equation} where each weight is non-negative $\beta_\ell \geq 0$ and summed to one: $\sum_{\ell=1}^{L} \beta_\ell = 1$.
OMA learns one scalar coefficient per model, shared across the
target adaptation set, by minimizing regularized predictive entropy.
All candidate-model parameters remain frozen during this step.
Once their predictive probabilities have been obtained, adaptation
operates on cached outputs: it requires neither access to
model internals nor backpropagation through the foundation models.
The number of optimized coefficients depends on the number of candidate models, rather than their parameter counts.

Predictive entropy provides an unlabeled adaptation signal, but
confidence is not, in general, equal to correctness. A model can make
sharply peaked yet incorrect predictions, and entropy minimization
can favour that model at the expense of both task performance and
calibration. We therefore make regularization central to OMA to balance closeness to a reference mixture and entropy reduction when determining the ensemble weights.
We examine its effect on calibration empirically: regularization
can reduce expected calibration error (ECE) in the evaluated
settings, but its benefits depend on the task and optimization
configuration. Since we have no access to target labels, regularization itself does not guarantee target-domain calibration.
Our evaluation therefore measures task performance and ECE
separately and examines how their trade-off relates to the learned
mixture composition.

We evaluate OMA on text and image classification tasks. Experiments on
four text classification tasks---MRPC, CoLA, RTE, and SST-2  ---
% ---across
% five random seeds characterize the effects of regularization
% strength, learning rate, optimization duration, and batch size on
% task performance, ECE, and model-mixing weights. 
% These experiments
identify settings with favourable performance--calibration
trade-offs.
% , while also revealing when the two metrics improve in
% different directions. 
%
Complementary image experiments on
ImageNet-1K \citep{deng2009imagenet} and five distribution-shift
benchmarks---ImageNet-V2, ImageNet-R, ImageNet-A, ImageNet-Sketch,
and ObjectNet---examine the potential of target-adaptive weighting
for zero-shot and lightly adapted visual models.

We designed our algorithm to be lightweight and accessible to researchers in all domains. Given that most common GPUs have under 24GB of memory due to cost, we chose our experiments to be runnable on a single \textit{NVIDIA RTX 4090} GPU with $24GB$ memory 
% (for loading large vision foundation models\footnote{\textit{EVA02-E-14-plus } model shown in \tabref{open_clip_models} requires $20.2$ GB memory to load the model on a GPU, signifying how memory-intensive fine-tuning the entire foundation models can be}) 
for all image classification experiments, and on a single \textit{NVIDIA RTX A4000} with $16GB$ memory for all text classification experiments. %
While ensembling model outputs increases inference-time computation, our method \textit{significantly reduces} \textbf{\textit{total}} \textit{computational cost} from fine-tuning to prediction, compared to fine-tuning the entire foundation model and then averaging the fine-tuned model weights. In the following section, we start by providing relevant background.

% Our computational objective is to reduce the cost of adapting an
% existing model pool, not to eliminate the inference cost of
% ensembling. The image experiments use a single NVIDIA RTX 4090 GPU
% with 24\,GB memory, and the text experiments use a single NVIDIA
% RTX A4000 GPU with 16\,GB memory. Although optimizing the mixture
% coefficients avoids further foundation-model fine-tuning,
% predicting new inputs still requires evaluating the participating
% models. We therefore distinguish adaptation cost from inference
% cost rather than claiming a universal reduction in total
% computation. OMA offers a practical mechanism for reusing
% heterogeneous predictors and incorporating additional models
% without retraining the existing candidates, while making the
% accuracy--calibration consequences of output adaptation explicit.

\section{Background}\label{sec:backgroud}

We first describe output averaging for heterogeneous classifiers, which serves as a baseline, and
clarify the distinction between predictive entropy, calibration, and source-informed model weighting.

\paragraph{Uniform output averaging.}
Consider $L$ frozen classifiers $\{M_\ell\}_{\ell=1}^{L}$ whose
predictions are probability distributions over the same label
space $\mathcal{Y}=\{1,\ldots,C\}$.  
% Write
% $p_\ell(y\mid\mathbf{x})$ for the predictive distribution of
% $M_\ell$, suppressing its fixed parameters.
% The classifiers may differ in architecture, pretraining data, and
% adaptation procedure; only their output label semantics and ordering
% must agree. 
% Model output averaging forms
% \begin{equation}
% \label{eq:predict}
%     q_{\boldsymbol{\beta}}(y\mid\mathbf{x})
%     = \sum_{\ell=1}^{L}\beta_\ell p_\ell(y\mid\mathbf{x}),
%     \qquad \boldsymbol{\beta}\in\Delta_L,
% \end{equation}
% where
% $\Delta_L=\{\boldsymbol{\beta}\in\mathbb{R}^{L}:
% \beta_\ell\geq0,\ \sum_{\ell=1}^{L}\beta_\ell=1\}$.
% These constraints ensure that the ensemble is a valid probability
% distribution. 
Uniform output averaging sets $\beta_\ell=1/L$ for an ensembled predictive distribution in \eqref{intro-oma-mixture}.
% OMA instead learns the weights from an unlabeled target set
% $\mathcal{U}_{\mathrm{T}}=\{\mathbf{x}_i\}_{i=1}^{M}$.
% A single weight vector is shared across that set, rather than
% predicting a separate vector for each input. 
% The component
% predictions need not be Bayesian posterior predictive, and no
In uniform output averaging, no common training dataset or architecture is required, making it easy to merge outputs from heterogeneous models. 

% \paragraph{Predictive entropy and calibration.}
% The average predictive entropy on the target inputs is
% \begin{equation}
% \label{eq:background-target-entropy}
%     \widehat{H}_{\mathrm{T}}(\boldsymbol{\beta})
%     = -\frac{1}{M}\sum_{i=1}^{M}\sum_{c=1}^{C}
%     q_{\boldsymbol{\beta}}(c\mid\mathbf{x}_i)
%     \log q_{\boldsymbol{\beta}}(c\mid\mathbf{x}_i),
% \end{equation}
% with $0\log0=0$. This quantity can be computed from cached
% probabilities without target labels. It measures uncertainty in
% the final predictive distribution, not correctness or an isolated
% epistemic-uncertainty. 

\paragraph{Expected calibration error
(ECE)} 
For a fixed ensemble $\vbeta$, we define a predicted label and confidence:
\begin{equation}
\label{eq:background-confidence}
\begin{aligned}
    \widehat{y}_{\boldsymbol{\beta}}(\mathbf{x}_i)
    &= \arg\max_{c\in\mathcal{Y}}
       q_{\boldsymbol{\beta}}(c\mid\mathbf{x}_i),\\
    s_{\boldsymbol{\beta}}(\mathbf{x}_i)
    &= \max_{c\in\mathcal{Y}}
       q_{\boldsymbol{\beta}}(c\mid\mathbf{x}_i).
\end{aligned}
\end{equation}
Confidence calibration asks whether predictions assigned confidence
$t$ are correct with frequency $t$. Expected calibration error
(ECE) estimates this discrepancy by grouping predictions into
confidence bins and comparing bin-wise accuracy and mean
confidence \citep{guo2017calibration}, for $n$ number of samples and $M$ number of bins:
\begin{align}
ECE &= \sum_{m=1}^M \frac{|B_m|}{n} \mid acc(B_m) - conf(B_m)\mid,
\end{align} 
where $acc(B_m) = \frac{1}{|B_m|}\sum_{i \in B_m} \mathbf{1}(\widehat{y}_{\boldsymbol{\beta}}(\mathbf{x}_i) == {y}_{true})$, and $conf(B_m)= \frac{1}{|B_m|}\sum_{i \in B_m} s_{\boldsymbol{\beta}}(\mathbf{x}_i).$ We use ECE to evaluate the calibration performance of ensembled models.

% Here, $n$ is the number of samples, $\hat{y}_i$ is a predicted label, $y_i$ is the true label, and $\hat{p}_i$ is the model's confidence for sample $i$, which is $\hat{p}_i = \max_c p(y_i=c|x_i)$.

% Entropy can decrease while calibration worsens, and
% source-domain calibration need not transfer under distribution
% shift \citep{ovadia2019trust}. We therefore distinguish label-free
% entropy optimization from calibration evaluation, which requires
% labels, and examine regularization as a control on adaptation
% rather than a calibration guarantee. We use ECE to evaluate the calibration performance of ensembled models. 

% Thus, the confidence used
% for ECE is $s_{\boldsymbol{\beta}}(\mathbf{x})$, not a model weight
% $\beta_\ell$. 
% ECE tells us how confident our model is by measuring how well the model's prediction accuracy matches the model's confidence about the prediction. Large ECE indicates the model's overconfidence, i.e., the model's prediction is confidently wrong. 

% In classification tasks, ECE is written as

\paragraph{Output ensembling in terms of logarithmic scores.}
Once an ensemble predictive distribution has been specified, what is the consequence of replacing it with a single component when evaluating predictions under the ensemble’s own predictive beliefs?
The following lemma states it in terms of \textit{logarithmic scoring rule}  \citep{Good1952}, 
% which assigns each event A that occurs a score of $-\log[p(A)]$.
% The negative logarithmic score 
evaluating a predictive distribution
$r$ by $-\log r(y)$ at the observed label $y$:
% \citep{Good1952}. The following identity clarifies a standard
% self-consistency property of model averaging without implying
% universal improvement over individual models.
\begin{mydefinitionbox}
\begin{lem}[Eq.4 in \citep{11c48783-68bb-36b7-8053-175211b0eaa8}]
\label{lemma:log_scoring_rule}
Fix an input $\mathbf{x}$ and weights
$\boldsymbol{\beta}\in\Delta_L$, and abbreviate
$q(c)=q_{\boldsymbol{\beta}}(c\mid\mathbf{x})$ and
$p_j(c)=p_j(c\mid\mathbf{x})$. For every $j\in\{1,\ldots,L\}$,
\begin{equation}
\label{eq:inequality}
    \mathbb{E}_{Y\sim q}[-\log q(Y)]
    \leq
    \mathbb{E}_{Y\sim q}[-\log p_j(Y)].
\end{equation}
If $p_j(c)=0$ for a class with $q(c)>0$, the right-hand side is
$+\infty$; otherwise both expectations are finite.
\end{lem}
\end{mydefinitionbox}
\noindent\textit{Proof.}
The difference between the right- and left-hand sides is
\begin{equation*}
    \sum_{c:q(c)>0}q(c)\log\frac{q(c)}{p_j(c)}
    = D_{\mathrm{KL}}(q\|p_j)\geq0. \qquad \square
\end{equation*}
% The proof states that the entropy of model-averaged prediction is always less than or equal to that of any single model's prediction. 
This identity compares the mixture with alternative predictions when outcomes are distributed according to the mixture itself. 
In the next section, we show that OMA directly optimizes the predictive entropy of model-averaged prediction as an unlabeled adaptation
surrogate, with an additional regularization. 

% It does not guarantee higher target accuracy, lower target ECE, or lower entropy than individual models.

% whose performance and calibration must be evaluated
% separately.

% The lemma compares the mixture's entropy $H(q)$ with the cross
% entropy $H(q,p_j)$, not with the component's entropy $H(p_j)$.
% Moreover, the expectation is under $q$, not the unknown target
% label distribution. It therefore guarantees neither better target
% accuracy nor lower target ECE. It also does not justify minimizing
% $H(q_{\boldsymbol{\beta}})$ across weight vectors, since the
% expectation distribution changes with $\boldsymbol{\beta}$.
% OMA instead uses predictive entropy as an unlabeled adaptation
% surrogate whose performance and calibration must be evaluated
% separately.

\paragraph{Source-informed reference weights.}
As a form of regularization, a reference vector $\boldsymbol{\beta}^{0}$ may be uniform or
constructed from source-domain information. Bayesian model
averaging (BMA) provides one such construction
\citep{11c48783-68bb-36b7-8053-175211b0eaa8}. Given labeled source
data $\mathcal{D}_{\mathrm{S}}$ and prior model probabilities
$p(M_\ell)$, the posterior model weights are
\begin{equation}
\label{eq:background-bma-weights}
    \beta_\ell^{0}
    = p(M_\ell\mid\mathcal{D}_{\mathrm{S}})
    = \frac{p(\mathcal{D}_{\mathrm{S}}\mid M_\ell)p(M_\ell)}
    {\sum_{k=1}^{L}p(\mathcal{D}_{\mathrm{S}}\mid M_k)p(M_k)},
\end{equation}
where the model evidence integrates its parameters:
\begin{equation}
\label{eq:background-evidence}
    p(\mathcal{D}_{\mathrm{S}}\mid M_\ell)
    = \int p(\mathcal{D}_{\mathrm{S}}\mid\mathbf{w}_\ell,M_\ell)
      p(\mathbf{w}_\ell\mid M_\ell)\,d\mathbf{w}_\ell.
\end{equation}
Exact BMA also integrates parameter uncertainty within each
component predictive distribution. Using fixed point-estimate
predictions and approximate evidence instead yields an
approximation to BMA. In our setting, this is an optional
source-side construction of reference weights, not a requirement
for probability-only target adaptation. Because the weights are
source-derived, they need not reflect target-domain model quality.

\section{Methods}\label{sec:methods}
% \section{Method}
% \label{sec:method}

We introduce \emph{Optimizable Model Averaging} (OMA), which adapts
output-merging weights to unlabeled target inputs by minimizing
regularized predictive entropy. The candidate models remain frozen;
only their contributions to the ensemble are optimized.

\subsection{Optimizable Model Averaging}
\label{subsec:OMA}

\paragraph{Setup.}
Let $\mathcal{U}_{\mathrm{T}}=\{\mathbf{x}_m\}_{m=1}^{M}$ be an
unlabeled target adaptation set, and let
$p_\ell(c\mid\mathbf{x})$ denote the probability assigned to class
$c\in\{1,\ldots,C\}$ by candidate model $M_\ell$.
The candidates may be zero-shot, fine-tuned, or MAP-based predictors
with different architectures, provided that their output classes
have the same semantics and ordering. Following the background,
OMA forms
\begin{equation}
\label{eq:conditional_likelihood_given_Dv}
    q_{\boldsymbol{\beta}}(c\mid\mathbf{x})
    = \sum_{\ell=1}^{L}\beta_\ell p_\ell(c\mid\mathbf{x}),
    \qquad \boldsymbol{\beta}\in\Delta_L,
\end{equation}
where $\Delta_L$ is the simplex of nonnegative weights summing to
one. A single coefficient vector is shared across the target set.
This is a weighted average of final predictive probabilities, not
an approximation obtained by combining the models' parameter
vectors. Once candidate predictions are cached, OMA requires no
access to model internals or gradients through the candidate models.

\paragraph{Motivation.}
Uniform averaging does not account for target-dependent differences
between models, while source-derived weights may become less
informative under distribution shift. Predictive entropy provides
a target-dependent signal that can be computed without labels:
concentrated predictions have lower entropy than diffuse ones.
This signal can be useful when prediction confidence reflects
correctness, but a confidently incorrect model can also have low
entropy. We therefore treat entropy as an adaptation surrogate,
not as a direct measure of target accuracy or calibration.

\paragraph{Predictive-entropy objective.}
For fixed candidate predictions, the empirical target entropy is
\begin{equation}
\label{eq:objective_not_final}
    \widehat{H}_{\mathrm{T}}(\boldsymbol{\beta})
    = -\frac{1}{M}\sum_{m=1}^{M}\sum_{c=1}^{C}
      q_{\boldsymbol{\beta}}(c\mid\mathbf{x}_m)
      \log q_{\boldsymbol{\beta}}(c\mid\mathbf{x}_m),
\end{equation}
with $0\log0=0$. The sum over classes evaluates the predictive
expectation exactly; neither target labels nor label sampling are
needed.

An important property of this objective motivates regularization.
Because entropy is concave and the mixture is linear in
$\boldsymbol{\beta}$,
\begin{equation}
\label{eq:oma-entropy-bound}
    \widehat{H}_{\mathrm{T}}(\boldsymbol{\beta})
    \geq \sum_{\ell=1}^{L}\beta_\ell
          \widehat{H}_{\mathrm{T}}(\mathbf{e}_\ell)
    \geq \min_{\ell}\widehat{H}_{\mathrm{T}}(\mathbf{e}_\ell),
\end{equation}
where $\mathbf{e}_\ell$ assigns all weight to model $\ell$.
The lower bound is attained by a minimum-entropy candidate, so the
unregularized problem over the simplex always admits a
single-model optimum. It does not inherently favour a nontrivial
ensemble, and the selected model need not be the most accurate or
best calibrated. If all candidates make identical predictive
distributions, the entropy is independent of the mixture weights.

\paragraph{Regularized OMA.}
We minimize a regularized objective over
$\boldsymbol{\beta}\in\Delta_L$ that constrains either the ensemble
predictions or the model weights:
\begin{equation}
\label{eq:objective}
    \boxed{
    \mathcal{J}_{\lambda}(\boldsymbol{\beta})
    = \widehat{H}_{\mathrm{T}}(\boldsymbol{\beta})
       +\lambda\mathcal{R}(\boldsymbol{\beta}),
    \qquad \lambda\geq0.
    }
\end{equation}
We consider the following two alternatives for $\mathcal{R}$.

\paragraph{(a) Prediction-space KL regularization.}
For each target input, construct a reference distribution from the
coordinate-wise median of the candidate probabilities:
\begin{equation}
\label{eq:oma-median-reference}
\begin{aligned}
    a_m(c) &= \operatorname{median}_{\ell=1,\ldots,L}
                    p_\ell(c\mid\mathbf{x}_m),\\
    r_m(c) &= \frac{a_m(c)}{\sum_{j=1}^{C}a_m(j)}.
\end{aligned}
\end{equation}
For even $L$, the median is the arithmetic mean of the two middle
values. The normalization is necessary because coordinatewise
medians need not sum to one. \eqref{oma-median-reference}
assumes a positive denominator, which holds for strictly positive
candidate probabilities. The reference is computed once from the
frozen predictions and remains fixed during optimization.
We define
\begin{equation}
\label{eq:oma-kl-regularizer}
\begin{aligned}
    \mathcal{R}_{\mathrm{KL}}(\boldsymbol{\beta})
    &= \frac{1}{M}\sum_{m=1}^{M}
       D_{\mathrm{KL}}\!\left(
          r_m\,\middle\|\,
          q_{\boldsymbol{\beta}}(\cdot\mid\mathbf{x}_m)
       \right)\\
    &= \frac{1}{M}\sum_{m=1}^{M}\sum_{c=1}^{C}
       r_m(c)\log
       \frac{r_m(c)}{q_{\boldsymbol{\beta}}(c\mid\mathbf{x}_m)}.
\end{aligned}
\end{equation}
This direction of KL penalizes assigning little probability to
classes supported by the reference. Unlike a source-derived
weight penalty, it uses only candidate predictions on the current
target inputs. Nevertheless, the median reference can inherit
shared model errors and is not assumed to be calibrated. It also
need not be representable by a single global mixture vector;
regularization encourages proximity, not exact agreement. We use this regularization in text classification.

\paragraph{(b) Weight-space $\ell_2$ regularization.}
Alternatively, given a fixed reference vector
$\boldsymbol{\beta}^{0}\in\Delta_L$, we use
\begin{equation}
\label{eq:oma-l2-regularizer}
    \mathcal{R}_{2}(\boldsymbol{\beta})
    = \lVert\boldsymbol{\beta}-\boldsymbol{\beta}^{0}\rVert_2^2.
\end{equation}
The reference can be uniform, approximate BMA weights, or weights
constructed from source-domain performance. Any performance-based
construction must yield nonnegative, normalized weights and must
not use target evaluation labels. This penalty limits departures
from the reference model composition, whereas
$\mathcal{R}_{\mathrm{KL}}$ constrains predictions directly.
In both cases, $\lambda$ controls the trade-off between entropy
reduction and the chosen regularizer. It anchors OMA to BMA only
in the weight-space variant when
$\boldsymbol{\beta}^{0}$ is set to BMA weights; it is not generally
an interpolation coefficient between BMA and OMA. We use this regularization in image classification.

\paragraph{Optimization.}
To enforce the simplex constraints, we optimize unconstrained
parameters $\boldsymbol{\tau}\in\mathbb{R}^{L}$ and set
\begin{equation}
\label{eq:oma-weight-parameterization}
    \alpha_\ell=\log(1+\exp(\tau_\ell)),
    \qquad
    \beta_\ell(\boldsymbol{\tau})
    =\frac{\alpha_\ell}{\sum_{k=1}^{L}\alpha_k}.
\end{equation}
Finite parameters give strictly positive weights summing to one;
boundary solutions can be approached in the limit.
Initializing $\boldsymbol{\tau}=\mathbf{0}$ gives uniform weights,
regardless of the regularization reference
$\boldsymbol{\beta}^{0}$. We apply gradient descent to
$\mathcal{J}_{\lambda}(\boldsymbol{\beta}(\boldsymbol{\tau}))$,
using the full cache or mini-batches of cached predictions.
For a mini-batch, the entropy and KL terms are averaged over its
examples, while the weight-space penalty remains unchanged.
The objective is not generally convex in $\boldsymbol{\beta}$, and the returned
$\widehat{\boldsymbol{\beta}}$ is a numerical solution. Algorithm~\ref{algo:oma} summarizes
the procedure.

\begin{algorithm}[t]
\caption{OMA with cached predictions}
\label{algo:oma}
\begin{algorithmic}[1]
\Require Cached probabilities
$P_{\ell mc}=p_\ell(c\mid\mathbf{x}_m)$;
regularizer choice; $\lambda\geq0$; $T$ steps;
batch size $B$; step sizes $\{\eta_t\}$;
$\boldsymbol{\beta}^{0}$ for weight-space regularization.
\Ensure Adapted weights $\widehat{\boldsymbol{\beta}}$ and
predictive distribution $q_{\widehat{\boldsymbol{\beta}}}$.
\If{prediction-space KL is used}
    \State Compute and fix $r_m$ for each input using
    \eqref{oma-median-reference}.
\EndIf
\State Initialize $\boldsymbol{\tau}^{(0)}\gets\mathbf{0}$.
\For{$t=0,\ldots,T-1$}
    \State Sample a batch
    $\mathcal{B}_t\subseteq\{1,\ldots,M\}$ of size $B$
    (or use all inputs).
    \State Form $\boldsymbol{\beta}(\boldsymbol{\tau}^{(t)})$
    using \eqref{oma-weight-parameterization} and compute
    the mixture probabilities on $\mathcal{B}_t$.
    \State Evaluate the batch version
    $\mathcal{J}_{\lambda,\mathcal{B}_t}$ of
    \eqref{objective} with the selected regularizer.
    \State Differentiate the batch loss through
    $\boldsymbol{\beta}(\boldsymbol{\tau})$ to obtain the
    gradient $\mathbf{g}_t$ with respect to $\boldsymbol{\tau}$.
    \State $\boldsymbol{\tau}^{(t+1)}\gets
    \boldsymbol{\tau}^{(t)}-\eta_t\mathbf{g}_t$.
\EndFor
\State Set $\widehat{\boldsymbol{\beta}}
\gets\boldsymbol{\beta}(\boldsymbol{\tau}^{(T)})$.
\State \Return $\widehat{\boldsymbol{\beta}}$ and
$q_{\widehat{\boldsymbol{\beta}}}(c\mid\mathbf{x})
=\sum_{\ell=1}^{L}\widehat{\beta}_\ell p_\ell(c\mid\mathbf{x})$.
\end{algorithmic}
\end{algorithm}

\paragraph{Computation and calibration.}
After caching, each optimization step requires
$O(BLC)$ arithmetic for a batch of $B$ inputs; storing the full
prediction cache requires $O(MLC)$ memory. These costs exclude
obtaining the candidate predictions. For new inputs, the learned
weights can be reused, but the participating candidate models must
still be evaluated. Neither regularizer directly optimizes ECE
or guarantees against overconfident errors. We therefore assess
task performance and calibration separately, using
$\max_c q_{\widehat{\boldsymbol{\beta}}}(c\mid\mathbf{x})$ as the
ensemble confidence. 

\section{Related Work}\label{sec:related_work}
% \section{Related Work}
% \label{sec:related-work}

We position OMA at the intersection of predictive ensembling and
unlabeled test-time adaptation, distinguishing methods by what they
combine, what information they require, and which quantities they adapt. Due to space limits, we describe only part of the relevant work here; \suppsecref{rest_related_work} describes the rest. 

\paragraph{Parameter-space model merging.}
Model Soups averages compatible fine-tuned checkpoints from
hyperparameter sweeps \citep{model_soups}, while Fisher-weighted
averaging incorporates parameter-importance information
\citep{matena2022fisher}. TIES-Merging addresses interference through
parameter trimming and sign selection \citep{yadav2023ties}, and
DARE sparsifies and rescales fine-tuning updates before merging
\citep{yu2024dare}. Particularly relevant is AdaMerging, which learns
task-wise or layer-wise parameter-merging coefficients by minimizing
entropy on unlabeled test inputs \citep{yang2024adamerging}.
These approaches combine corresponding parameters of compatible
models. OMA instead combines final predictive probabilities, so its
candidates need not share an architecture or initialization.
This flexibility retains the inference cost of evaluating the
participating models, whereas parameter merging produces a single
model for deployment.

\paragraph{Test-time ensemble-coefficient learning.}
Several methods studied test-time coefficient learning. DECISION jointly adapts source feature extractors and
aggregation weights through conditional-entropy minimization,
marginal-diversity maximization, and pseudo-labelling
\citep{ahmed2021decision}. 
Importantly, its
\emph{DECISION-weights} ablation freezes the feature extractors and
optimizes only aggregation weights; its pooling precedes the
softmax rather than averaging final probabilities. 
EMEA learns
sentence-specific combinations of intermediate language-adapter
representations through predictive-entropy minimization
\citep{wang2021emea}. SADE learns weights for frozen, deliberately
specialized experts by maximizing prediction stability across
augmented target views \citep{zhang2022sade}. CEW combines
validation-informed class weighting with prediction-entropy-based
test-time ensembling \citep{song2024cew}.
These are direct precedents for unlabeled coefficient learning.
% OMA focuses on a shared arithmetic 
While we share the test-time coefficient learning, our focus is the investigation of a particular 
% regularized probability-mixture formulation
mixture of heterogeneous
predictive distributions, with regularization on either its target
predictions or its model weights. 
We additionally study the calibration and accuracy trade-offs under different regularization strengths.
% The relevant distinctions are the
% pooling operator, regularizer, and required information, rather than
% entropy minimization or frozen-model weighting alone.

\section{Experiments}\label{sec:experiments}

\begin{figure*}[htbp]
    \centering
    \includegraphics[width=1\textwidth]{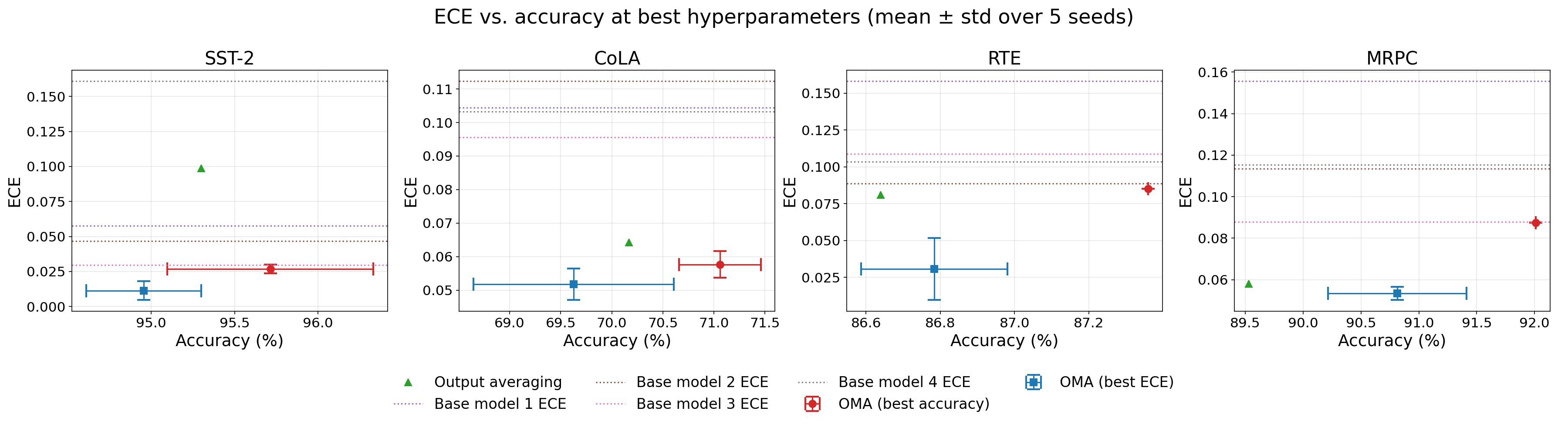}
    \caption{ECE versus Accuracy}
    \label{fig:ece_vs_acc_text}
\end{figure*}

\begin{table}[htbp]
\centering
\caption[Accuracy and ECE of output averaging and OMA]{Accuracy\protect\footnote{Following \citep{model_soups}, we mean Accuracy as the average of classification accuracy and $F_1$ score for MRPC, classification accuracy for RTE and SST-2, and Matthews correlation for CoLA \citep{matthews1975comparison}.} (\%) and ECE of output averaging and OMA at the best-accuracy and best-ECE hyperparameter settings (mean $\pm$ std over 5 seeds)\protect\footnote{These configurations were selected retrospectively using evaluation labels and characterize the attainable performance–calibration trade-off within the search grid. They are not used as evidence for label-free hyperparameter selection.}. Best values per dataset are in bold.}
\label{tab:best_results_text}
\scalebox{0.7}{\begin{tabular}{llcc}
\toprule
Dataset & Method & Accuracy ($\uparrow$) & ECE ($\downarrow$) \\
\midrule
\multirow{3}{*}{SST-2}
 & Output averaging      & 95.30            & 0.099 \\
 & OMA (best accuracy)   & \textbf{95.71} $\pm$ 0.62 & 0.027 $\pm$ 0.003 \\
 & OMA (best ECE)        & 94.95 $\pm$ 0.35 & \textbf{0.011} $\pm$ 0.007 \\
\midrule
\multirow{3}{*}{CoLA}
 & Output averaging      & 70.17            & 0.064 \\
 & OMA (best accuracy)   & \textbf{71.06} $\pm$ 0.40 & 0.058 $\pm$ 0.004 \\
 & OMA (best ECE)        & 69.63 $\pm$ 0.98 & \textbf{0.052} $\pm$ 0.005 \\
\midrule
\multirow{3}{*}{RTE}
 & Output averaging      & 86.64            & 0.081 \\
 & OMA (best accuracy)   & \textbf{87.36} $\pm$ 0.00 & 0.085 $\pm$ 0.000 \\
 & OMA (best ECE)        & 86.78 $\pm$ 0.20 & \textbf{0.031} $\pm$ 0.021 \\
\midrule
\multirow{3}{*}{MRPC}
 & Output averaging      & 89.53            & 0.058 \\
 & OMA (best accuracy)   & \textbf{92.01} $\pm$ 0.00 & 0.088 $\pm$ 0.000 \\
 & OMA (best ECE)        & 90.81 $\pm$ 0.60 & \textbf{0.053} $\pm$ 0.003 \\
\bottomrule
\end{tabular}}
\end{table}

\begin{table}[t]
    \centering
\scalebox{0.7}{
\begin{tabular}{c|cccc}
\toprule
\multicolumn{1}{l|}{}              & MRPC                 & RTE                  & CoLA                 & SST-2                \\ \hline
bert-base                         & 86.44                & 65.34                & 59.05                & 92.43                \\
bert-large                        & 90.61                & 61.73                & 64.55                & 93.58                \\
funnel (xlarge-base)       & 92.01                & 87                   & 69.6                 & 95.99                \\
funnel (xlarge)            & 89.69                & 87                   & 68.06                & 54.93                \\ \hline
% \multicolumn{1}{l}{}              & \multicolumn{1}{l}{} & \multicolumn{1}{l}{} & \multicolumn{1}{l}{} & \multicolumn{1}{l}{} \\
% output avg                               & 89.32                & 87.18                & 70.6                 & 95.29                \\
% \rowcolor[HTML]{00FFFF} 
% OMA                               & \textbf{92.10}       & \textbf{88.62}       & \textbf{72.59}       & \textbf{96.62}       \\  
 
                             % & \rowcolor[HTML]{00FFFF}\textbf{(+3.11\%)}       & \textbf{(+1.66\%)}       & \textbf{(+2.81\%)}       & \textbf{(+1.4\%)}       \\ \hline 
% \hline 
% \multicolumn{1}{l}{}              & \multicolumn{1}{l}{} & \multicolumn{1}{l}{} & \multicolumn{1}{l}{} & \multicolumn{1}{l}{} \\
MS (Bert, best) &88.3&	61	&59.1	&92.5          \\
MS (Bert, greedy) & 88.3                 & 61.7                   & 59.1                 & 93                   \\
MS (T5, best)         & 91.8	&78.3	&58.8	&94.6              \\
MS (T5, greedy)           & 92.4                 & 79.1                 & 60.2                 & 94.7                 \\
Fisher(Bert)               & 85.1                 & 73.2                 & 55.8                 & 92.4    \\ \hline   
\end{tabular}}
\caption{Performance of other methods on GLUE benchmark. \textbf{Top}: individual model performance when fine-tuned on each dataset. \textbf{Bottom}: Performance of existing methods. Model soups (MS) on two different architectures (Bert-base-uncased and T5-base \citep{raffel2020t5}). Fisher merging (Fisher) on Bert-base architecture.}
\label{tab:comparison_OMA}
\end{table}

We consider two classification tasks. The first is text classification, where we ensemble the outputs of fine-tuned transformer-based models. 
The second is image classification, where we freeze the pre-trained OpenCLIP models, fine-tune an additional linear classifier on ImageNet-1K data, and then ensemble the outputs of the fine-tuned linear classifiers and zero-shot models.

\subsection{Text Classification Tasks}

% Preamble
% \usepackage{booktabs}

\begin{figure}[t]
\centering
  % \centering
\includegraphics[width=0.9\textwidth]{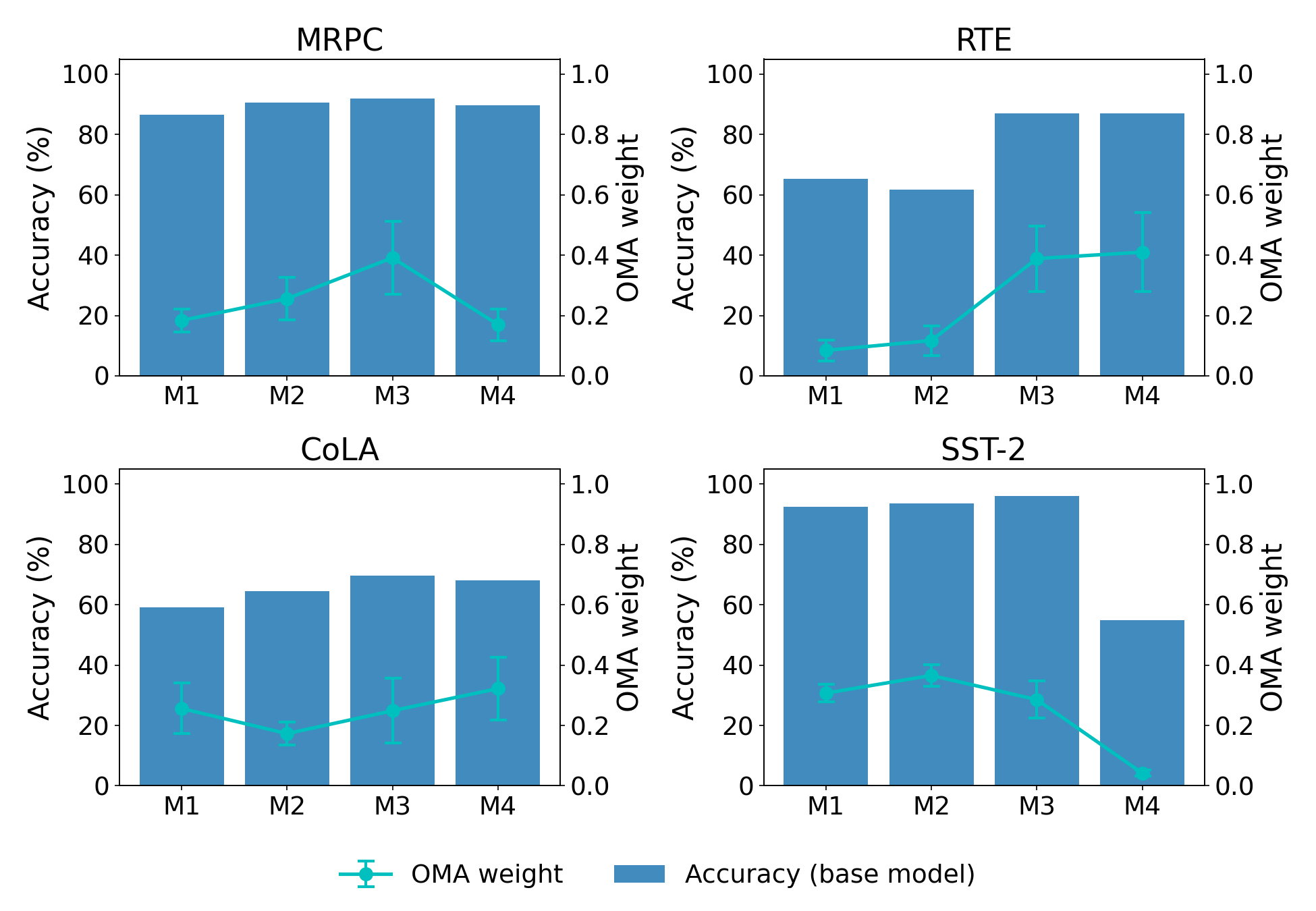}
% }{%
\caption{OMA weights at the best ECE, learned by \algoref{oma} and individual model's performance
}\label{fig:oma_weight}
\end{figure}
% \end{table*}

\begin{table}[t]
\scalebox{0.7}{
\begin{tabular}{c|cccc}
\multicolumn{1}{l|}{}              & MRPC                 & RTE                  & CoLA                 & SST-2                \\ \hline
% Model	MRPC	RTE	COLA	SST-2
BERT-Base	&74.80	&53.07	&43.57	&88.19 \\
BERT-Large&	74.98	&50.90	&53.91	&91.63\\
funnel(xlarge-base)	&74.38&	52.71	&13.44&	92.89\\
funnel(xlarge)	&74.80	&49.10	&42.68	&94.50\\
  \hline  
LoRA (output avg) &74.80&	59.21	&51.48	&92.54\\
Bayesian LoRA &74.98&	50.90&	53.91&	91.74\\
 \hline
\end{tabular}}
\caption{Performance of LoRA and Bayesian LoRA on GLUE benchmark. \textbf{Top}: individual model performance when fine-tuned on each dataset using LoRA. \textbf{Bottom}: Performance of output averaging and Bayesian LoRA}
\label{tab:comparison_BayesianLoRA}
\end{table}

Following \citep{model_soups}, we consider four text classification tasks from the GLUE benchmark \citep{wang2018glue}: MRPC \citep{dolan2005automatically}, RTE \citep{dagan2005pascal, bar2006second, giampiccolo2007third, bentivogli2009fifth}, CoLA \citep{warstadt2018neural} and SST-2 \citep{socher2013recursive}, as in \citep{dodge2020fine}. Details of these datasets are given in \suppsecref{text_classification}. 
% Following \citep{model_soups}, we use the average of classification accuracy and $F_1$ score for MRPC, classification accuracy for RTE and SST-2, and Matthews correlation for CoLA \citep{matthews1975comparison}.
%

For each dataset, we fine-tune four different models: two BERT-base cased models (large and base) \citep{devlin-etal-2019-bert} and two funnel-transformers (xlarge and xlarge-base) \citep{funnel_transformer}. We downloaded the pre-trained models from \url{https://huggingface.co/models}. See the list of checkpoints and ablation study in \suppsecref{checkpoints} and how we perform text classification in these models in \suppsecref{bert_classification}.  In OMA, we used the KL regularizer in \eqref{oma-kl-regularizer}.

\tabref{best_results_text} shows that OMA achieves a better accuracy-calibration trade-off than uniform averaging. \figref{ece_vs_acc_text} shows ECE against accuracy for OMA at its best-accuracy and best-ECE hyperparameter settings (mean ± std over 5 seeds), alongside output averaging and the four base models' ECE. Across all four datasets, OMA tuned for ECE is better calibrated than output averaging and every base model. An ablation study over different hyperparameter settings is given in \suppsecref{ablation_text}.

\tabref{comparison_OMA} shows OMA outperforms output averaging and a single best model. 
 % For Model Soups, we took the numbers from Table 5 of \citep{model_soups}. For Fisher merging (last row), we took the results from Table A1  of \citep{Fisher_Merging}, the best value across different rows under each column.
Since Model Soups and Fisher merging consider different architectures than ours, directly comparing ours to their results is unreasonable. What matters is the improvement over each best model, which is larger in our case than in Model Soups. 
We also show the learned model weights $\vbeta$ matching the best ECE in \tabref{best_results_text} in \figref{oma_weight}. Interestingly, the learned model weights do not behave similarly to each model's performance.  That is, models with high accuracy do not necessarily get higher weights, and vice versa. 

\tabref{comparison_BayesianLoRA} shows the performance of individually trained models with LoRA on the GLUE dataset, as well as that of output averaging and Bayesian LoRA. In LoRA training, we used the same hyperparameters for all models as shown in \suppsecref{details_lora}. 
Surprisingly, the Bayesian LoRA approach is not consistently better than simple output averaging. The reason for this could be that we were constrained to use a diagonal Hessian for the LoRA parameters of the \textit{last layer only} due to memory limitations. Note that the number of LoRA parameters of these models ranges between 200,000 and 1,000,000, and the last layer's LoRA parameters alone range between 20,000 and 50,000, for which computing the full Hessian was not feasible
% We instead computed the diagonal Hessian for computational tractability
on the GPU we had.

\begin{table*}[t]
\centering
\caption{Top-1 accuracy (\%) of output averaging and OMA for different candidate model pools. Numbers in parentheses are the relative improvement of OMA over output averaging; the best accuracy in each column is in bold. These numbers are highest accuracy from \figref{ece_vs_acc_img}.}
% \label{tab:main_results}
\label{tab:image-ours}
\scalebox{0.75}{
\begin{tabular}{|l|l|c|c|c|c|c|c|}
\cline{3-8}
\multicolumn{2}{l|}{} & \cellcolor{headgray}Img-1K & \cellcolor{headgray}Img-V2 & \cellcolor{headgray}Img-R & \cellcolor{headgray}Img-sketch & \cellcolor{headgray}Img-A & \cellcolor{headgray}ObjNet \\
\hline
\rowcolor{poolyellow} & Output avg & 84.90 & 79.14 & 96.22 & 75.85 & 85.08 & 77.13 \\
\cline{2-8}
\rowcolor{poolyellow} & OMA & 85.11 & 78.91 & 96.17 & \textbf{76.23} & 86.68 & 77.64 \\
\rowcolor{poolyellow} \multirow{-3}{*}{Zeroshot} & & (+0.25\%) & (-0.29\%) & (-0.05\%) & (+0.50\%) & (+1.88\%) & (+0.66\%) \\
\hline
\rowcolor{poolpink} & Output avg & 88.98 & 81.53 & 94.36 & 72.76 & 83.61 & 73.90 \\
\cline{2-8}
\rowcolor{poolpink} & OMA & \textbf{89.19} & 81.90 & 94.64 & 72.68 & 85.47 & 74.02 \\
\rowcolor{poolpink} \multirow{-3}{*}{MAP} & & (+0.24\%) & (+0.45\%) & (+0.30\%) & (-0.11\%) & (+2.22\%) & (+0.16\%) \\
\hline
\rowcolor{poolblue} & Output avg & 88.15 & 81.73 & 96.09 & 75.84 & 86.19 & 76.78 \\
\cline{2-8}
\rowcolor{poolblue} & OMA & 88.48 & \textbf{81.94} & \textbf{96.37} & 75.76 & \textbf{88.04} & \textbf{78.39} \\
\rowcolor{poolblue} \multirow{-3}{*}{Zeroshot\&MAP} & & (+0.37\%) & (+0.26\%) & (+0.29\%) & (-0.11\%) & (+2.15\%) & (+2.10\%) \\
\hline
\end{tabular}}
\end{table*}

\subsection{Image Classification Tasks}

% \mj{Describe how this part is different from the text classification cases. Explain BMA a bit here.}

For image classification tasks, we consider eight\footnote{ We provide a detailed explanation of the OpenCLIP repository in \suppsecref{openclip_model_overview}. \tabref{open_clip_models} shows our chosen models (named as fe-1, $\cdots$, fe-8), based on average zero-shot performance and training-data diversity.} pre-trained OpenCLIP vision foundation models \citep{ilharco_gabriel_2021_5143773}. 
Because these trained models provide rich visual features and fine-tuning them is computationally expensive, we use them as fixed feature extractors and add a trainable linear classifier to each.  

By restricting the linear classifiers to be the only trainable parameters, we can form Bayesian model averaging (BMA) straightforwardly: We find the MAP estimate (given in \eqref{map}) of the eight classifiers using the training data of Imagenet-1K. We then calculate the posterior model weights using Bayesian model averaging (BMA), as in \eqref{marginal3} and \eqref{marginal4}. We provide detailed explanations of BMA in \suppsecref{BMA_full}. 

 We use these posterior model weights as a fixed reference vector $\vbeta^0$, and evaluate how OMA using the L2-regularizer in \eqref{oma-l2-regularizer} performs on Imagenet-1K and five out-of-distribution datasets: Imagenet-V2 \citep{imagenetv2}, Imagenet-A \citep{imagenet-a}, Imagenet-Sketch \citep{imagenet-sketch}, Objectnet \citep{objectnet}, Imagenet-R \citep{imagenet-r}.  For the candidate model pool, we consider the eight MAP estimates trained on Imagenet-1K and eight zero-shot models from the chosen OpenCLIP models (i.e., 16 candidate models). 

While  OMA does not require Bayesian candidate models or posterior
inference, the use of BMA (freezing pretrained feature extractors, training lightweight linear classifiers, and approximating their model
evidence to obtain the reference mixture) allows us to study how source-informed
weighting and target entropy interact, under distribution
shift.

% We first describe an open-source repository called \textit{OpenCLIP} \citep{ilharco_gabriel_2021_5143773}, from which we took pre-trained vision foundation models.

Additional results for 
Camelyon17 \citep{camelyon}, 
Sun397 \citep{Xiao2010SUNDL}, and 
Flowers102 \citep{Nilsback08},  are in \suppsecref{Other_Data}. 
%

% Unlike BMA,  OMA does not necessitate introducing and fine-tuning linear classifiers.

% Note that OMA does not require Bayesian candidate models or posterior
% inference. When labelled source data and feature access are
% available, we additionally use approximate Bayesian model averaging
% (BMA) to construct source-informed reference weights for image
% classification. We freeze pretrained visual feature extractors,
% train lightweight linear classifiers, and approximate their model
% evidence to obtain the reference mixture. This source-side
% construction is optional and distinct from OMA's probability-only
% target adaptation. It allows us to study how source-informed
% weighting and target entropy interact without treating the
% source-derived weights as necessarily reliable under distribution
% shift.

% \suppsecref{BMA_full}

\tabref{image-ours} shows that OMA combining the MAP estimates performs the best 
in the in-distribution Img-1K validation data.
The domain shift from ImageNet-1K to ObjectNet, A, Sketch, and R is arguably large. So, OMA combining the MAP estimates as well as 
incorporating knowledge from the zero-shot models was useful for those datasets, except for Sketch, where OMA combining zero-shot models alone outperforms other pools.

% in the in-distribution Img-1K validation data.
Ablation studies with different hyperaparameter values are given in \suppsecref{ablation}.
% combining the MAP estimates and zero-shot classifiers improves the performance of BMA with MAP estimates with a large margin for ObjectNet, Img-A, Imag-Sketch, and Img-R. 
The individual models' performance for each dataset is shown in \suppsecref{individual}, where the zero-shot models for Img-A, Sketch and Objectnet outperform those fine-tuned for Img-1K, indicating these datasets are indeed quite different from Img-1K.  
See \suppsecref{whywhy} for our explanations on why BMA’s source-data–determined weights become unreliable under distribution shift, and why OMA’s test-time entropy minimization avoids this failure mode.

\tabref{image_comparison_table} shows simple baselines using the same candidate pools, such as output merging with each model's accuracy as the weight and averaging outputs from a single base model with different initializations. 
We also include Model soups \citep{model_soups}, CoCa  \citep{yu2022coca}
and EVA02-L \citep{EVA02L} (See \suppsecref{comparison}) for context.

% ?, as direct comparison is not sensible which differ in architecture, training data, and fine-tuning protocol, so we include them for context, rather than as direct comparisons.
% We also add 
% The estimated posterior model weights are visualized in \figref{model_posterior_weight}, where fe-1 and fe-4 are deemed more useful than others for Imagenet-1K. 

% % \vspace{-0.2cm}
% % \begin{wrapfigure}{r}{0.5\textwidth}
% \begin{figure}[t]
%   \begin{center}
%    \centering\includegraphics[width=0.8\textwidth]{imagenet.pdf}
%   \end{center}
%   \caption{Model posterior weights given the Imagenet-1K training data. 
% The first and the fourth feature extractors are deemed the most significant.
% }
% \label{fig:model_posterior_weight}
% % \end{wrapfigure}
% \end{figure}

% If BMA + OMA is still better, it would greatly improve the significance of the paper. In addition, the paper claims that OMA works well when there is a strong distribution shift. However, it’s not very clear from the experiment that it is the case where the performance of OMA is fairly similar to the output average. Would it be possible to consider a dataset with a very strong distribution shift to see if it really helps in that case? Finally, the paper claims that the method is more computationally efficient than the baseline, such as model soup. I think it makes sense. However, it would be nice to see the exact runtime even for a small number of data points and compare with a non-Bayesian baseline as well.

\begin{table}[t]
\centering
\small
\caption{\textbf{TOP:} Other baselines using the same candidate models in \tabref{image-ours}. \textbf{Bottom:} External reference numbers from prior methods. These methods differ in architecture, training data, and fine-tuning protocol, so we include them for context, rather than as direct comparisons. Accuracy values are percentages. 
% Bold indicates the best, and underline indicates the second best. 
% Model Soups (greedy) by \citet{pmlr-v162-wortsman22a}, CoCa (frozen) by \citet{yu2022coca}, 
 % and EVA-02-L by \citet{EVA02L}
}
\label{tab:image-external}
\scalebox{0.75}{
\begin{tabular}{lrrrrrr}
\toprule
Method & Img-1K & V2 & R & Sketch & A & ObjNet \\
% \midrule
% Uniform avg & 88.97 & 81.53 & 94.32 & {75.48} & 86.1 & 76.6 \\
\midrule
Accuracy-weighted & 88.98 & 81.83 & 94.38 & 73.95 & 83.89 & 74.05 \\
\midrule
Single best model & 88.27 & 80.30 & {95.80} & 74.66 & 86.06 & 76.42 \\
\midrule
\midrule 
Model Soups, greedy  & {90.94} & {84.22} & 95.46 & 74.23 & {92.67} & {78.52} \\
\midrule
CoCa, frozen & {90.6} & -- & -- & -- & -- & -- \\
\midrule
EVA-02-L & 90.0 & {82.4} & 89.9 & 70.1 & 87.7 & 62.8 \\
\bottomrule
\end{tabular}}
\label{tab:image_comparison_table}
\end{table}
\begin{figure*}[htbp]
    \centering
    \includegraphics[width=1\textwidth]{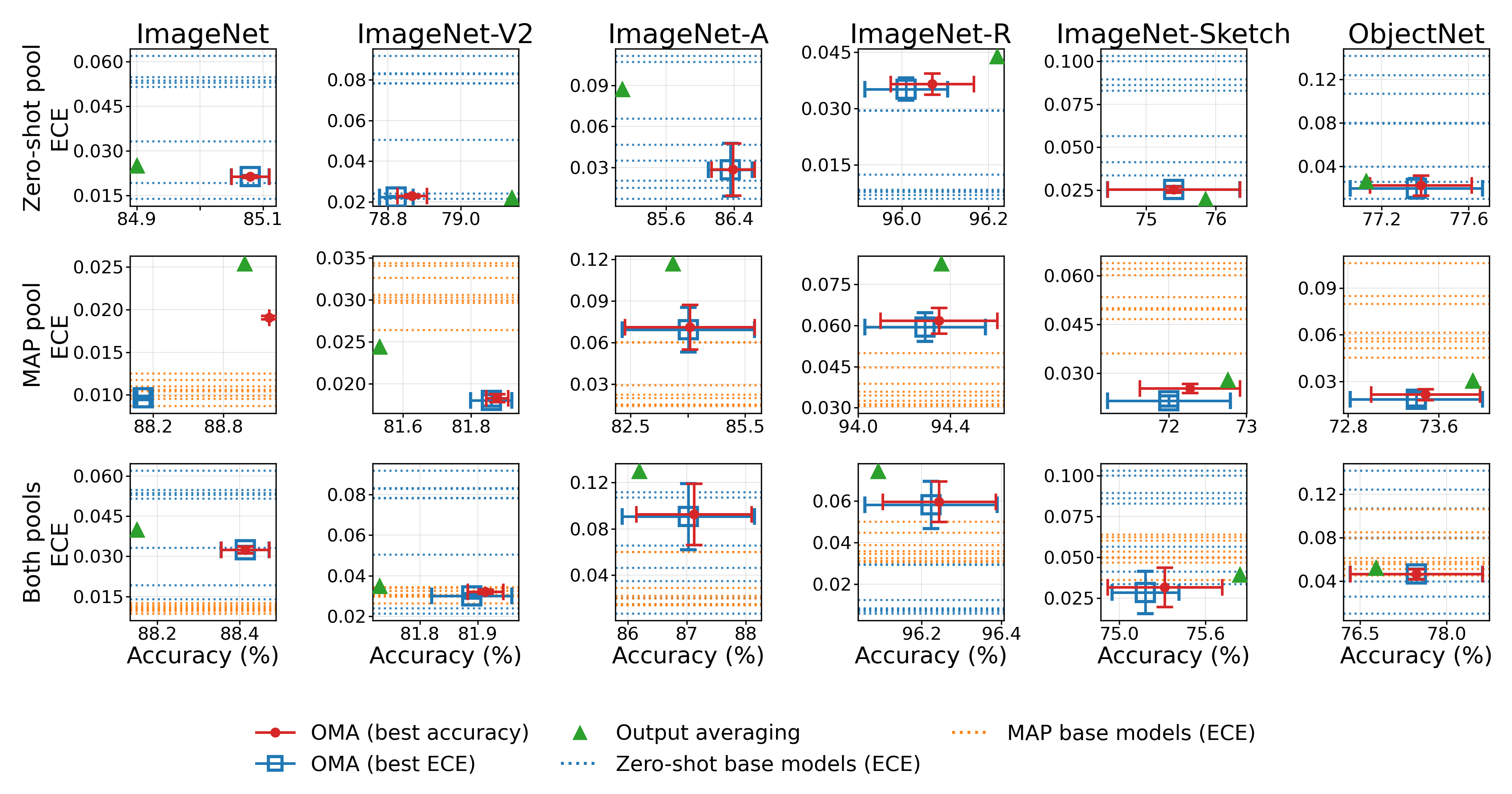}
    \caption{ECE versus Accuracy across six datasets (each column) and three candidate pools (each row) (error bars showing the standard deviation over seeds)}
    \label{fig:ece_vs_acc_img}
\end{figure*}

\subsection{Required computation}
Output merging, like ours, yields more inference-time costs than parameter merging. However, our framework requires significantly less \textit{total} computation, as described in \suppsecref{required_computation_bma_vs_others}. In summary, fine-tuning a single model in Model Soup takes 3 days on an A100, while BMA on an NVIDIA RTX 4090 GPU (worse than an A100) needs $86.1\%$ lower than Model Soup. When comparing fine-tuning all 58 models in Model Soup versus 8 models in BMA, Model Soup takes 4176 hours, and BMA takes 80 hours, yielding a $98.08\%$ decrease in total runtime. However, the models' accuracy is comparable: Model Soup on ImageNet-1K achieves $90.94$, and our OMA using BMA weights achieves $89.19$.

% \mj{Add ECE results here: explain how to read \figref{ece_vs_acc_img}.}

\paragraph{ECE vs Accuracy}
\figref{ece_vs_acc_img} shows the trade-offs between ECE and accuracy under three model pools: zero-shot models, MAP estimates, and both. Each column shows the three cases for each dataset. \figref{lambda_ablation_image} shows an ablation study with $\lambda$.

% Across all datasets, the ECE of MAP estimates (orange dotted lines) is significantly lower than the ECE of zero-shot models (blue dotted lines), indicating that MAP estimates are better calibrated than zero-shot models, although calibration depends on the dataset.
% Since MAP estimates are trained with ImageNet-1K data, those MAP estimates' ECE (second row) gets worse as domain shift gets larger (e.g., ObjectNet, Sketch, and A have larger ECE than ImageNet-1K and V2)

% the three pools sit on very different ECE scales: zero-shot base models are badly calibrated (ECE about 0.65–0.95) and MAP base models are well calibrated (about 0.01–0.11). The ensembles follow the pool they are built from. 

% For ImageNet-1K and V2, using both pools achieves better accuracy and ECE than using either pool.  

% Figure: ECE vs. accuracy of OMA at its best-accuracy and best-ECE settings, output averaging, and the individual zero-shot and MAP base models (ECE only), . 

Combining only zero-shot models through OMA (first row) reduces ECE and achieves better accuracy than uniform merging on ImageNet, A, and Sketch, while uniform merging achieves higher accuracy on V2 and R; however, OMA has lower ECE on R.
Combining only MAP estimates through OMA (second row) is better calibrated than uniform merging on ImageNet, V2, A, and R; while uniform merging achieves better accuracy on Sketch and ObjectNet.
Combining both pools through OMA yields generally better ECE and accuracy than uniform merging, except on Sketch, where OMA's ECE is lower but its accuracy is worse than uniform merging.

% , so the two selection criteria agree in most panels. The clear exceptions are ImageNet-R, where the MAP pool's best-accuracy configuration has ECE near 0.94 against 0.07 for its best-ECE one, and the ImageNet-A both-pool panel, where the two criteria pick different settings. 

% zError bars are wide in some panels (ObjectNet, ImageNet-A) and several points rest on only two seeds, so small differences should be read with caution.

% Because the linear classifiers were trained with ImageNet-1K, 

\section{Summary and Discussion}\label{sec:summary_discussion}
We developed OMA to improve text classification performance by learning model-ensemble weights via regularized entropy minimization.
We explored BMA and OMA to merge foundation model outputs by training only linear classifiers for image classification. 
We also examined trade-offs between accuracy and calibration at varying regularization strengths. 
The inference cost of large foundation models can be burdensome, and all output-merging methods suffer from this; remedies include quantization and model pruning described in \suppsecref{remedy}. 
 % If there is a strict restriction on inference time, one could choose to use weight merging schemes instead. However, one cannot merge the weights of different architectures. Each scheme has its own benefits. Users are free to choose which one to use, based on when a particular scheme is more sensible and efficient to use.
%
Our ultimate goal is to develop computationally efficient methods by utilizing pre-trained or lightly fine-tuned models, reducing the need for additional energy-intensive training. 
Such methods seem timely and the right and responsible way forward.

\newpage

\section*{AI use statement}

We used generative AI tools, to assist with refining research hypotheses and experimental designs, reviewing mathematical derivations and proof arguments, analyzing and interpreting experimental results, and implementing and debugging experimental code. We also used these tools to draft and revise portions of the manuscript, improve clarity and organization, prepare LaTeX pseudocode, and identify and discuss relevant literature. 

The authors reviewed and revised the AI-assisted material included in this work. We checked mathematical statements and derivations, verified references against the cited sources, and checked reported numerical claims against experimental outputs. AI-generated code was reviewed and revised by the authors before use in the experiments. The authors made the final methodological decisions and determined the interpretation and presentation of the findings. We take responsibility for the final content of this work, including all text, claims, and artifacts produced with the assistance of generative AI.

\section*{Acknowledgements}
We thank our anonymous reviewers for their constructive feedback, which has helped significantly improve
our manuscript. M. Park was supported in part by the Natural Sciences and Engineering Research Council of
Canada (NSERC) and the Canada CIFAR AI Chairs program. M. Park was also funded by Novo Nordisk
Fonden RECUIT grant no.0065800 during her stay at the Technical University of Denmark, where the initial idea was developed.

\bibliography{reference}
% Bibliography style is set once to apalike in the preamble.

\clearpage
\appendix
\thispagestyle{empty}

% Supplementary material: To improve readability, you must use a single-column format for the supplementary material.
\onecolumn
\aistatstitle{Supplementary Materials}

\section{Bayesian model averaging}\label{supp:BMA_full}
% % add a schematic of how embeddings are combined for linear classifier part

\paragraph{Notation} We denote the pre-trained and frozen $l$-th open-clip model (which will use as a feature extractor) by $\vphi_l$, where $l \in \{1, 2, \cdots, L\}$. We also denote the linear classifier for $l$-th feature extractor by $\vw_l$, where $\vw_l \in \mathbb{R}^{CD}$, $C$ is the number of classes, and $D$ is the dimension of features.  
The $l$-th linear classifier's parameters are denoted by $\vw_l$ and the $l$-th frozen feature extractor's parameters are denoted by $\vtheta_l$. Hence, the parameters of the $l$-th model are the union of both parameters $\{\vtheta_l, \vw_l\}$.
Below, we describe essential quantities for performing BMA\footnote{We point curious readers to \suppsecref{thoughtsBMA} about our thoughts on BMA versus mixture-of-experts (ME) formulation.}.

\subsubsection{A Challenge in computing the marginal likelihood of each model}
Here we treat the parameters of the linear classifier $\vw_l$ to be random variables and fix the parameters of the feature extractor to the pre-trained values denoted by $\hat\vtheta_l$. We write the marginal likelihood of data for model $M_l$ as: 
\begin{align}\label{eq:marginal}
    &p(\Dat | M_l) \nonumber \\
    % &= \int p(\Dat | \vtheta_k, M_k) p(\vtheta_k|M_k) d\vtheta_k. \\
    &= \int_{\vw_l, \vtheta_l} p(\Dat | \vtheta_l, \vw_l, M_l) p(\vw_l|\vtheta_l, M_l) p(\vtheta_l|M_l) d\vw_l d\vtheta_l, \nonumber \\
    &\approx \int_{\vw_l} p(\Dat | \hat\vtheta_l, \vw_l, M_l) p(\vw_l|\hat\vtheta_l, M_l)  \delta_{\vtheta_l= \hat\vtheta_l}d\vw_l, \nonumber \\
    &\approx \int_{\vw_l} p(\Dat |\vw_l, M_l) p(\vw_l|M_l) d\vw_l,
\end{align} where, in the last line, we drop the dependency on $\hat\vtheta_l$ for notational simplicity.
Under this formulation, we now specify the likelihood term: $p(\Dat |\vw_l, M_l)$ and the prior term: $p(\vw_l|M_l) $. 

\paragraph{Likelihood}
In the multi-class classification, $p(\Dat |\vw_l, M_l)$ is often assumed to be a categorical distribution, where $\vy_n^{c}$ is a vector of length $C$ where only one entry is $y_n^{c}=1$ while the rest entries are $0$,   
\begin{align}\label{eq:likelihood}
    p(\Dat |\vw_l, M_l) &= \prod_{n=1}^N \prod_{c=1}^C p(y_n^{c}|\vphi_l(\vx_n),\vw_l, M_l), \nonumber \\
    &= \scalemath{0.9}{\prod_{n=1}^N \prod_{c=1}^C \left[\frac{\exp(\vphi_l(\vx_n)^T \vw_l^{c})}{\sum_{c'=1}^C \exp(\vphi_l(\vx_n)^T \vw_l^{c'})}\right]^{y_n^{c}}}. 
\end{align} 
Notice that this is the likelihood of the multi-class logistic regression where the inputs are the features from the foundation models. This becomes handy when we compute the Hessian matrix later. 

\paragraph{Prior}
We put a Gaussian prior with varying precision on each dimension of the weights:
\begin{align}
    p(\vw_l|\alpha) &= \prod_{i=1}^D \prod_{j=1}^C \Nrm \left(w_{l, ij} \;|\; 0, \; \alpha \right) = \Nrm(\vw_l\;|\;\mathbf{0}, \; S_\valpha),\nonumber 
\end{align} 
where the vectorized notation has a diagonal covariance matrix $S_\alpha = \diag[{\alpha}\bm{I}]$ with a hyperparameter $\alpha$.

\paragraph{The challenge} Under the prior and likelihood terms described above, computing the integral in \eqref{marginal} to obtain the marginal likelihood of each model is analytically intractable. If we had small models with only tens of parameters, the integral could be computed numerically using Monte Carlo integration. 
However, for datasets like imagenet-1K and the aforementioned CLIP models, $D\approx 10^3$ and $C=1000$, we have more than $10^6$ parameters where sampling-based approaches perform poorly. 

We adopt the \textit{Laplace approximation} \citep{pauler_bayes_1999}, which approximates the posterior as a multivariate Gaussian distribution for computational tractability. See \suppsecref{comparison_MC}, where we run some low-dimensional experiments to compare the runtime and accuracy of sampling-based approaches with our Hessian-based approach described next.

\subsubsection{Laplace approximation}
In Laplace approximation, 
we consider the unnormalized log-posterior given by:
\begin{align}\label{eq:un_norm_log_post}
    \Psi(\vw_l) 
    % &= \log p(\Dat|\vw_l) + \log p(\vw_l|\alpha), \nonumber\\
    &= \scalemath{0.92}{\sum_{n=1}^N \sum_{c=1}^C y_n^{c} \left[\vphi_l(\vx_n)\tr \vw_l^{c} - 
    \log(\sum_{c'=1}^C \exp(\vphi_l(\vx_n)\tr\vw_l^{c'})) \right]} \nonumber \\
    & \quad- \tfrac{1}{2}\vw_l\tr S_{\alpha}^{-1} \; \vw_l  
     - \tfrac{1}{2} \log| 2 \pi S_\alpha|.
\end{align}
We then approximate $\Psi(\vw_l) $ as a quadratic function in $\vw_l$ using the 2nd-order Taylor expansion:
\begin{align}\label{eq:2ndorder_approx}
    \hat\Psi(\vw_l) &\approx \Psi(\vw_{l}^{map}) \nonumber \\
    &\quad - \tfrac{1}{2}({\vw_l} - \vw_{l}^{map})^T \Sigma^{-1} ({\vw_l}- \vw_{l}^{map}).
    % \nonumber 
\end{align}
The first derivative yields the maximum a posteriori (MAP) estimate of $\vw_l$ and the second derivative yields the Hessian: 
\begin{align}\label{eq:map}
    \tfrac{\partial}{\partial \vw_l}\Psi(\vw_l) = 0
    % nonumber \\
    % &= \frac{\partial}{\partial \vw_l} \log p(\Dat|\vw_l) - \tfrac{1}{2}S_{\alpha}^{-1} \; \vw_l = 0
    \; \mapsto \;  \vw_{l}^{map}, &     \nonumber \\
    \underbrace{- \tfrac{\partial^2}{\partial \vw_l \partial \vw_l\tr}\log p(\Dat|\vw_l)|_{\vw_l =  \vw_{l}^{map}}}_{:=H}
 + S_{\alpha}^{-1}
   \; &:=  \; \Sigma^{-1}. 
\end{align} 
 Under our formulation, as this formulation is essentially equivalent to that of multi-class Logistic Regression, there exists a closed-form expression (See \suppsecref{hessian_comput}) for the full Hessian.
However, the Hessian matrix $H$ is $DC \times DC$, roughly $10^6 \times 10^6$, and loading such a matrix on a GPU with limited memory is infeasible. In this paper, we adopt the block-diagonal Hessian approach. Using a block diagonal Hessian is a sensible choice, as the diagonal blocks dominate the off-diagonal blocks under the multi-class logistic regression models in Appendix G.1 of \citep{Neurips2024_350e718f}.
\suppsecref{hessian_comput} shows that the diagonal elements are orders of magnitude larger than the off-diagonal elements in the subset of the full Hessian.

\paragraph{Approximate model marginal likelihood}

With these quantities above, 
the log marginal likelihood simplifies to 
\begin{align}\label{eq:marginal3}
\log p(\Dat|M_l) &\approx \log p(\Dat|\vw_{l}^{map}, M_l) 
     - \tfrac{1}{2}\vw_{l}^{map}\tr S_\alpha^{-1} \; \vw_{l}^{map} \nonumber \\
     & \quad - \tfrac{1}{2} \log |H S_\alpha + I|.
\end{align} 

% \vspace{-0.5cm}
\subsubsection{Posterior predictive distribution}

The approximate marginal likelihood in \eqref{marginal3} gives us an un-normalized posterior model probability if we assume a uniform distribution for $p(M_l)=\frac{1}{L}$. Hence, we normalize it by dividing the sum
\begin{align}\label{eq:marginal4}
    p(M_l | \Dat) & = \frac{p(\Dat | M_l) }{\sum_{l'=1}^L p(\Dat|M_{l'})}.
\end{align}

To compute the posterior predictive distribution given in \eqref{predict_approx}, we need to compute $p(\vy^*|\vx^*, M_l, \Dat)$. 
We approximately compute this term: 
\begin{align}\label{eq:conditional_likelihood}
    p(\vy^*|\vx^*, M_l, \Dat)
    &= \int p(\vy^*|\vx^*, \vw_l, M_l, \Dat)
    p(\vw_l| M_l, \Dat)d {\vw_l} \nonumber \\
    &\approx  \int  p(\vy^*|\vx^*, \vw_l, M_l, \Dat) \delta_{\vw_l=\vw_{map}} d {\vw_l} \nonumber \\
    &\approx p(\vy^*|\vx^*,\vw_{l}^{map}, M_l, \Dat).
\end{align} 
The posterior predictive distribution under BMA is
\begin{mydefinitionbox}
\begin{align}\label{eq:predict_approx}
    p(\vy^*|\vx^*, \Dat)= \sum_{l=1}^L p(\vy^*|\vx^*, \vw_{l}^{map}, M_l, \Dat) p(M_l | \Dat).
\end{align}
\end{mydefinitionbox}
% where the last line is due to the Laplace approximation. 
The BMA algorithm is given in Algorithm \ref{algo:BMA}.

\section{OpenClip models}\label{supp:openclip_model_overview} 

\paragraph{OpenClip}
Among the many publicly available vision foundation models, we focus on OpenClip models in this paper. OpenClip is an open-source implementation of OpenAI's CLIP (Contrastive Language-Image Pre-training). OpenClip models' architectures and training datasets are publicly available, unlike those of OpenAI's CLIP models. 
The repository provides an extensive list of CLIP models at varying architectures and sizes, with the datasets used for training those models, such as LAION-400M, LAION-2B and DataComp-1B. In total, the repository contains $121$ combinations of CLIP models and training datasets. See \url{https://github.com/mlfoundations/open_clip} for detailed descriptions of each model architecture and dataset.
The OpenClip repository includes zero-shot performance of each model evaluated on $38$ different datasets\footnote{\url{https://github.com/mlfoundations/open_clip/blob/main/docs/openclip_results.csv}}. Among them, we choose \textit{top} eight open-clip models as feature extractors, where chosen feature extractors are shown in \tabref{open_clip_models} (named as fe-1, $\cdots$, fe-8). The choice is based on the average zero-shot performance and the diversity of training data.

We evaluated the performance of our method on commonly used image classification datasets, including Imagenet-1k\footnote{\url{https://huggingface.co/datasets/imagenet-1k}} \citep{deng2009imagenet}, 
and five Imagenet out-of-distribution datasets: Imagenet-V2 \citep{imagenetv2}, Imagenet-A \citep{imagenet-a}, Imagenet-Sketch \citep{imagenet-sketch}, Objectnet \citep{objectnet}, Imagenet-R \citep{imagenet-r}. 
We consider all eight feature extractors described in \tabref{open_clip_models}.

\begin{table*}[t]
\centering
\rowcolors{2}{gray!10}{white}
\scalebox{0.67}{\begin{tabular}{@{}l l l r r r c@{}}
\toprule
\textbf{Name in Paper} & \textbf{OpenCLIP Name} & \textbf{Pretraining Data} & \textbf{Params (M)} & \textbf{FLOPs (B)} & \textbf{Memory (G)} & \textbf{Avg. Perf. (38 sets)} \\
\midrule
\textcolor{blue}{fe-1} & ViT-H-14-378-quickgelu         & dfn5b                   & 986.71  & 1054.05 & 4.4   & 0.7079 \\
\textcolor{blue}{fe-2} & ViT-H-14-quickgelu             & dfn5b                   & 986.11  & 381.68  & —     & 0.6961 \\
\textcolor{blue}{fe-3} & \textit{EVA02-E-14-plus}       & \textit{laion2b\_s9b\_b144k} & \textit{5044.89} & \textit{2362.19} & \textit{20.2} & \textit{0.6930} \\
\textcolor{blue}{fe-4} & ViT-SO400M-14-SigLIP-384       & webli                   & 877.96  & 723.48  & 4.1   & 0.6921 \\
\textcolor{blue}{fe-5} & ViT-bigG-14-CLIPA-336          & datacomp1b              & 2517.76 & 2271.58 & 10.35 & 0.6842 \\
—                      & ViT-bigG-14-CLIPA              & datacomp1b              & 2517.22 & 1007.93 & —     & 0.6822 \\
—                      & ViT-SO400M-14-SigLIP           & webli                   & 877.36  & 233.54  & —     & 0.6808 \\
\textcolor{blue}{fe-6} & EVA02-E-14                     & laion2b\_s4b\_b115k     & 4704.59 & 2311.42 & 18.8  & 0.6690 \\
—                      & ViT-L-14-quickgelu             & dfn2b                   & 427.62  & 175.33  & —     & 0.6687 \\
—                      & ViT-L-16-SigLIP-384            & webli                   & 652.48  & 422.91  & —     & 0.6683 \\
—                      & ViT-H-14-CLIPA-336             & datacomp1b              & 968.64  & 800.88  & —     & 0.6677 \\
\textcolor{blue}{fe-7} & ViT-H-14-quickgelu             & metaclip\_fullcc        & 986.11  & 381.68  & 4.4   & 0.6671 \\
% \vdots & \vdots            & \vdots      & \vdots  & \vdots  & \vdots   & \vdots 
% \\
\textcolor{blue}{fe-8}& Convnext\_xxlarge
& laion2b\_s34b\_b82k\_augreg\_soup & 1200.58& 443.03 & — & 0.6530 \\
\bottomrule
\end{tabular}}
\caption{We select $8$ pre-trained open-clip models \citep{ilharco_gabriel_2021_5143773} based on their performance of zero-shot classifiers evaluated on $38$ datasets and the diversity of the pretraining datasets. }
% We call each selected open-clip model by fe-1, fe-2, $\cdots$, fe-7 (in blue).}
\label{tab:open_clip_models}
\end{table*}

% \section{Algorithms}

% \usepackage{algorithmic}
\begin{algorithm}[h]
% \scalebox{0.8}{
  \caption{BMA for VFMs}
  \label{algo:BMA}
\begin{algorithmic}
  \State {\bfseries Input:} 
  Pretrained feature extractors: $\vphi_1, \cdots, \vphi_L$. Training data $\Dat$ and validation data $\Dat_v$. 
  \State {\bfseries Output: Posterior predictive on $\Dat_v$} 
  \State \textbf{Step 1.} \textit{Pre-Process Data:} Take feature representations of both training and validation datasets, $\Dat$ and $\Dat_v$, by feedforwarding each datapoints through each feature extractors.
  \State \textbf{Step 2.} \textit{Model posterior weights:}
\For{$l=1$ {\bfseries to} $L$}
  \State \textbf{1} Compute MAP estimate and Hessian given \eqref{map}  using $\Dat$. 
  \State  \textbf{2} Compute unnormalized model posterior weight given in \eqref{marginal3}   using $\Dat$. 
\EndFor
\State \textbf{Step 3.} \textit{Posterior predictive:}  Normalize the posterior model weights using \eqref{marginal4} and compute the predictive distribution given in \eqref{conditional_likelihood} for $\Dat_v$ 
\end{algorithmic}
% }
\end{algorithm}

% \begin{algorithm}[h]
% % \scalebox{0.8}{
%   \caption{OMA for FMs}
%   \label{algo:OMA}
% \begin{algorithmic}
%   \STATE {\bfseries Input:} 
%   Models $\{M_l\}_{l=1}^L$, validation data $\Dat_v$, training steps $P$, a contatnt $\lambda$, and prior $\beta_l^{0}$.
%   \STATE {\bfseries Output:  Predictive distribution on $\Dat_v$} 
%   \STATE \textbf{Step 1.} \textit{
%   Train for the model weights $\beta_{l}$:} Optimize the weights by minimizing \eqref{objective}. 
% \STATE \textbf{Step 2.} \textit{Posterior predictive:}  Compute the predictive distribution given in \eqref{conditional_likelihood_given_Dv} for $\Dat_v$ 
% \end{algorithmic}
% % }
% \end{algorithm}

\section{A few thoughts on BMA}\label{supp:thoughtsBMA}

\subsection{BMA Versus ME}

One might wonder why we use BMA (Bayesian model averaging) as opposed to ME (mixture of experts).  Chapter 14 in \citep{bishop} says that \textit{in Bayesian model averaging, the whole data set is assumed to be generated by a single model. By contrast, in the model combination like in ME, different data points within the data set can potentially be generated from different values of the latent variable, i.e., different components.}

    In the Bayesian sense, as we gather more data, eventually one model will be better than others as the posterior over that model will be peaked. Do we expect such a phenomenon in this framework? The `model posterior weights' tell us which model is more useful than others for classifying a query point, but because we are fixing the feature extractor $\vphi_l$ to the pre-trained values, we do not think the posterior over the model will be peaked necessarily (although the posterior over $\vw_l$ will be peaked as we have more training data). 
     
     In that sense, conceptually, this framework seems closer to ME than BMA. However, we do not necessarily assume each query point is coming from a different model. 
     %
     % We assume each point goes through one model, which is a collection of VFMs we consider. So, it is also not true that we assume each query point is coming from only a single VFM among the collection of models we consider.
     %
     Furthermore, what distinguishes this method from ME is that we do not use the expectation maximization (EM) algorithm to fit the model. We use the Laplace approximation to get the model posterior weight.

     Overall, we do not claim BMA is more suitable than ME for the model merging problem. These come from different viewpoints, and choosing which one to prefer can be problem-specific. In this particular case of merging predictions from trained foundation models, these two might give a similar performance. Nevertheless, it is an intriguing future direction to test ME in the same setting.

\subsection{Some thoughts on different dimensionality of weights across different foundation models}
One might wonder if it makes sense to apply BMA to different model sizes. We assume that 
 each of the candidate models belongs to some model class. We assume this model class is defined by a \textit{super-set} model, which encompasses all connections and architectures of candidate (or subset) models. So, each candidate model can be viewed as a super-set model with different zero-padded connections (if the candidate does not contain such connections that exist in the super-set model). This is similar to the case of linear models in the classical BMA settings, where a different selection of linear weights is considered a candidate model.
 % \vfill

%      \section{Code}
% Our code is given in the anonymous repo: \url{https://anonymous.4open.science/r/Code_BMA4VFMs-FE07}

 \section{Minibatch Hessian computation}\label{supp:hessian_comput}
 
 we will consider the unnormalized log-posterior given by:
\begin{align}
    &\Psi(\vw_l) = \log p(\Dat|\vw_l) + \log p(\vw_l|\alpha), \nonumber\\
    &= \scalemath{0.95}{\sum_{n=1}^N \sum_{c=1}^C y_n^{c} \left[\vphi_l(\vx_n)\tr \vw_l^{c} - 
    \log(\sum_{c'=1}^C \exp(\vphi_l(\vx_n)\tr\vw_l^{c'})) \right]} \nonumber \\
    & \quad - \tfrac{1}{2}\vw_l\tr S_{\alpha}^{-1} \; \vw_l  
     - \tfrac{1}{2} \log| 2 \pi S_\alpha|.
\end{align}
We want to approximate $\Psi(\vw_l) $ as a quadratic function in $\vw_l$ using the 2nd-order Taylor expansion.
\begin{align}
    \hat\Psi(\vw_l) &\approx \Psi(\vw_{l}^{map}) - \tfrac{1}{2}({\vw_l} - \vw_{l}^{map})^T \Sigma^{-1} ({\vw_l}- \vw_{l}^{map}) \nonumber 
\end{align}
where the first derivative of this equation gives us the maximum a posteriori estimate of $\vw_l$
\begin{align}
    \frac{\partial}{\partial \vw_l}\Psi(\vw_l)
    &= \frac{\partial}{\partial \vw_l} \log p(\Dat|\vw_l) - S_{\alpha}^{-1} \; \vw_l = 0
    \; \mapsto \;  \vw_{l}^{map},  
\end{align} and the second derivative gives us the  posterior precision matrix  
\begin{align}
    &- \frac{\partial^2}{\partial \vw_l \partial \vw_l\tr}\Psi(\vw_l) |_{\vw_l =  \vw_{l}^{map}} \nonumber \\
    &= 
    \underbrace{- \frac{\partial^2}{\partial \vw_l \partial \vw_l\tr}\log p(\Dat|\vw_l)|_{\vw_l =  \vw_{l}^{map}}}_{:=H}
 + S_{\alpha}^{-1}
   \; :=  \; \Sigma^{-1}. \nonumber
    \end{align}
The size of the Hessian matrix  $H$ is $DC \times DC$. For datasets like imagenet-1K, this becomes approximately $10^6 \times 10^6$, which is prohibitive. We resort to the block diagonal of Hessian instead. For each class, we consider a $D \times D$ Hessian matrix and compute the eigen-values of each block diagonal (since the eigenvalues of a block diagonal matrix are the concatenation of the eigenvalues of each block).

To make the computation quicker, we also subsample the datapoints when computing the eigenvalues of each block diagonal Hessian.
\begin{align}
H &= - \frac{\partial^2}{\partial \vw_l \partial \vw_l\tr}\log p(\Dat|\vw_l)|_{\vw_l =  \vw_{l}^{map}} \\
&\approx - \frac{\partial^2}{\partial \vw_l \partial \vw_l\tr} \left[\frac{N}{M}\sum_{n=1}^M \sum_{c=1}^C y_n^{c} \left[\vphi_l(\vx_n)\tr \vw_l^{c} - 
    \log(\sum_{c'=1}^C \exp(\vphi_l(\vx_n)\tr\vw_l^{c'})) \right] \right]|_{\vw_l =  \vw_{l}^{map}}
\end{align}

The block-diagonal term of the Hessian matrix wrt $\vw_k$ (for class $k$) is given by:
\begin{align}
H_{kk}= \nabla_{\mathbf{w}_k}^2 \mathcal{L}(\mathbf{W}) = \frac{N}{M} \sum_{i=1}^M p(\mathbf{x}_i)_k (1 - p(\mathbf{x}_i)_k) \mathbf{x}_i \mathbf{x}_i^\top 
\end{align}
where $p(\vx_i)_k = \sigma(W_{map} \vx_i)$ and $\sigma$ is the softmax function. 

The off-diagonal terms of the Hessian matrix (for $k \neq k'$) are given by 
\begin{align}
H_{kk'} = \nabla_{\mathbf{w}_k}\nabla_{\mathbf{w}_{k'}}\mathcal{L}(\mathbf{W}) = \frac{N}{M} \sum_{i=1}^M p(\mathbf{x}_i)_k (- p(\mathbf{x}_{i})_{k'}) \mathbf{x}_i \mathbf{x}_i^\top 
\end{align}

% Note that both $\vw_{\text{map}}$ is a function of $\valpha$ and $H$ is also a function of $\valpha$ because we evaluated the Hessian at $\vw_{\text{map}}$.
% Now the problem boils down to finding a posterior distribution over the weights in each model $p(\vw_k | \hat\vphi_k, \Dat)$.

\paragraph{Approximate model marginal likelihood}

% The log marginal likelihood is
% \begin{align}
% \log p(\Dat|\vw_{l}^{map}) 
%      - \tfrac{1}{2}\vw_{l}^{map}\tr S_\alpha^{-1} \; \vw_{l}^{map} - \tfrac{1}{2} \log |H S_\alpha + I|
% \end{align} 
When we use the mini-batch Hessian, the first term in \eqref{marginal3} is approximated by
\begin{align}
    \log p(\Dat|\vw_{l}^{map}) \approx \frac{N}{M}\sum_{n=1}^M \sum_{c=1}^C y_n^{c} \left[\vphi_l(\vx_n)\tr \vw_l^{c} - 
    \log(\sum_{c'=1}^C \exp(\vphi_l(\vx_n)\tr\vw_l^{c'})) \right]|_{\vw_l =  \vw_{l}^{map}}
\end{align}
The last term in \eqref{marginal3} can be written as
\begin{align}
    \tfrac{1}{2} \log |H S_\alpha + I|= \tfrac{1}{2} \sum_{j=1}^J \log(\lambda_j \alpha + 1)
\end{align}

Once we compute this quantity, we divide it by the number of training datapoints, then exponentiate it. So that this value is comparable to datasets of different sizes.

The following plot shows the subset of Hessian under each feature extractor with its corresponding MAP estimate trained for ImageNet-1K. For traceability, we subsampled 20 classes and 20 feature dimensions, yielding the [400 by 400] subset of the full Hessian.   
\begin{figure}[ht!]
    \centering

    % First row
    \begin{subfigure}[b]{0.32\textwidth}
        \includegraphics[width=\textwidth]{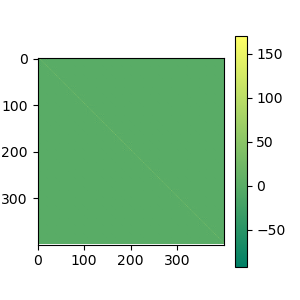}
        \caption{Hessian with fe-1}
    \end{subfigure}
    \hfill
    \begin{subfigure}[b]{0.32\textwidth}
        \includegraphics[width=\textwidth]{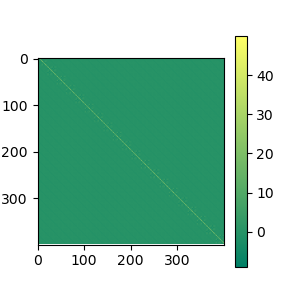}
        \caption{Hessian with fe-2}
    \end{subfigure}
    \hfill
    \begin{subfigure}[b]{0.32\textwidth}
        \includegraphics[width=\textwidth]{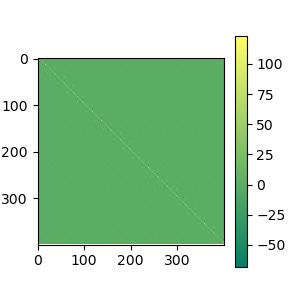}
        \caption{Hessian with fe-3}
    \end{subfigure}

    % \vspace{0.5cm}

    % Second row
    \begin{subfigure}[b]{0.32\textwidth}
        \includegraphics[width=\textwidth]{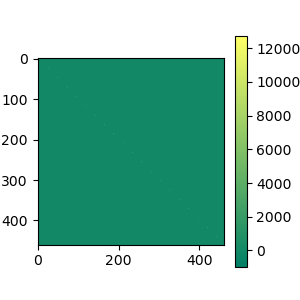}
        \caption{Hessian with fe-4}
    \end{subfigure}
    \hfill
    \begin{subfigure}[b]{0.32\textwidth}
        \includegraphics[width=\textwidth]{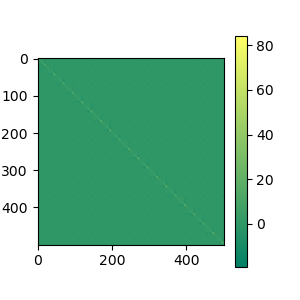}
        \caption{Hessian with fe-5}
    \end{subfigure}
    \hfill
    \begin{subfigure}[b]{0.32\textwidth}
        \includegraphics[width=\textwidth]{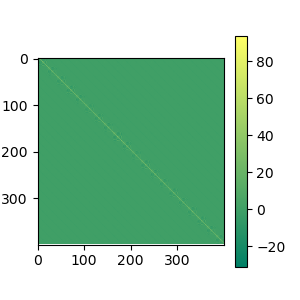}
        \caption{Hessian with fe-6}
    \end{subfigure}

    % \vspace{0.5cm}

    % Third row (single plot centered)
    \begin{subfigure}[b]{0.32\textwidth}
        \centering
        \includegraphics[width=\textwidth]{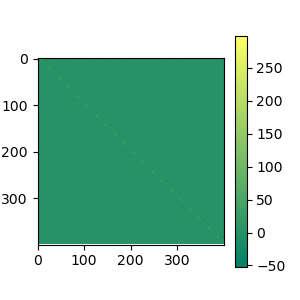}
        \caption{Hessian with fe-7}
    \end{subfigure}
    \begin{subfigure}[b]{0.32\textwidth}
        \centering
        \includegraphics[width=\textwidth]{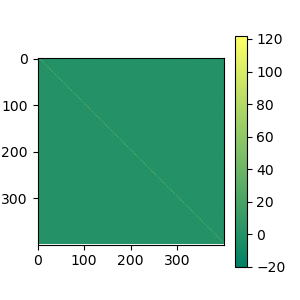}
        \caption{Hessian with fe-8}
    \end{subfigure}

    \caption{The diagonal elements are orders of magnitude larger than off-diagonal elements (with the MAP estimates trained for ImageNet-1K. Performance in \tabref{image-ours}).
Showing the magnitude of a subset of the Hessian for a [400 × 400]
subset of the Hessian, which is originally roughly [1 million by 1 million], sampling 20 classes log-uniformly and 20 input dimensions uniformly.}
    \label{fig:seven-plots}
\end{figure}

% \paragraph{When are the diagonal blocks orders of magnitude larger than the off diagonal blocks in the Hessian under our model?}

% Recall that 
% the block-diagonal term of the Hessian matrix wrt $\vw_k$ (for class $k$) is
% \begin{align}
% H_{kk}= \nabla_{\mathbf{w}_k}^2 \mathcal{L}(\mathbf{W}) = \frac{N}{M} \sum_{i=1}^M p(\mathbf{x}_i)_k (1 - p(\mathbf{x}_i)_k) \mathbf{x}_i \mathbf{x}_i^\top 
% \end{align}
% where $p(\vx_i)_k = \sigma(W_{map} \vx_i)$ and $\sigma$ is the softmax function. 
% And the off-diagonal terms of the Hessian matrix (for $k \neq k'$) are 
% \begin{align}
% H_{kk'} = \nabla_{\mathbf{w}_k}\nabla_{\mathbf{w}_{k'}}\mathcal{L}(\mathbf{W}) = \frac{N}{M} \sum_{i=1}^M p(\mathbf{x}_i)_k (- p(\mathbf{x}_{i})_{k'}) \mathbf{x}_i \mathbf{x}_i^\top 
% \end{align}

% This means, when our model is confident

\section{Performance of individual models for ImageNet-1K and its variant datasets}\label{supp:individual}

% In the main text, we listed only seven foundation models used, while the experiments for ImageNet-1K and its variants were done using eight foundation models (sorry for this mistake!). 
The full list is given in 
\tabref{open_clip_models_correct}. 
\begin{table}[h]
\centering
\rowcolors{2}{gray!10}{white}
\scalebox{0.67}{\begin{tabular}{@{}l l l r r r c@{}}
\toprule
\textbf{Name in Paper} & \textbf{OpenCLIP Name} & \textbf{Pretraining Data} & \textbf{Params (M)} & \textbf{FLOPs (B)} & \textbf{Memory (G)} & \textbf{Avg. Perf. (38 sets)} \\
\midrule
\textcolor{blue}{fe-1} & ViT-H-14-378-quickgelu         & dfn5b                   & 986.71  & 1054.05 & 4.4   & 0.7079 \\
\textcolor{blue}{fe-2} & ViT-H-14-quickgelu             & dfn5b                   & 986.11  & 381.68  & —     & 0.6961 \\
\textcolor{blue}{fe-3} & \textit{EVA02-E-14-plus}       & \textit{laion2b\_s9b\_b144k} & \textit{5044.89} & \textit{2362.19} & \textit{20.2} & \textit{0.6930} \\
\textcolor{blue}{fe-4} & ViT-SO400M-14-SigLIP-384       & webli                   & 877.96  & 723.48  & 4.1   & 0.6921 \\
\textcolor{blue}{fe-5} & ViT-bigG-14-CLIPA-336          & datacomp1b              & 2517.76 & 2271.58 & 10.35 & 0.6842 \\
—                      & ViT-bigG-14-CLIPA              & datacomp1b              & 2517.22 & 1007.93 & —     & 0.6822 \\
—                      & ViT-SO400M-14-SigLIP           & webli                   & 877.36  & 233.54  & —     & 0.6808 \\
\textcolor{blue}{fe-6} & EVA02-E-14                     & laion2b\_s4b\_b115k     & 4704.59 & 2311.42 & 18.8  & 0.6690 \\
—                      & ViT-L-14-quickgelu             & dfn2b                   & 427.62  & 175.33  & —     & 0.6687 \\
—                      & ViT-L-16-SigLIP-384            & webli                   & 652.48  & 422.91  & —     & 0.6683 \\
—                      & ViT-H-14-CLIPA-336             & datacomp1b              & 968.64  & 800.88  & —     & 0.6677 \\
\textcolor{blue}{fe-7} & ViT-H-14-quickgelu             & metaclip\_fullcc        & 986.11  & 381.68  & 4.4   & 0.6671 \\
% \vdots & \vdots            & \vdots      & \vdots  & \vdots  & \vdots   & \vdots 
% \\
\textcolor{blue}{fe-8}& Convnext\_xxlarge
& laion2b\_s34b\_b82k\_augreg\_soup & 1200.58& 443.03 & — & 0.6530
\\
\bottomrule
\end{tabular}}
\caption{We select $8$ pre-trained open-clip models \citep{ilharco_gabriel_2021_5143773} based on their performance of zero-shot classifiers evaluated on $38$ datasets and the diversity of the pretraining datasets. }
% We call each selected open-clip model by fe-1, fe-2, $\cdots$, fe-7 (in blue).}
\label{tab:open_clip_models_correct}
\end{table}
We show the individual model's performance on Imagenet and its variant datasets in \tabref{individual_imagenet}.

\begin{table}[h]
\centering
\begin{tabular}{|l|l|r|r|r|r|r|r|r|r|}
\hline
                &          & \multicolumn{1}{l|}{fe-1} & \multicolumn{1}{l|}{fe-2} & \multicolumn{1}{l|}{fe-3} & \multicolumn{1}{l|}{fe-4} & \multicolumn{1}{l|}{fe-5} & \multicolumn{1}{l|}{fe-6} & \multicolumn{1}{l|}{fe-7} & \multicolumn{1}{l|}{fe-8} \\ 
                \hline
imagenet        & map      & 88.05                    & 87.11                    & 87.98                    & 88.15                    & 88.27                    & 87.86                    & 86.06                    & 85.49                    \\
                & zeroshot & 84.51                    & 83.57                    & 82.52                    & 83.05                    & 83.19                    & 82.1                     & 80.64                    & 78.92                    \\ \hline
imagenet-v2     & map      & 80.29                    & 78.89                    & 80.12                    & 80.31                    & 80.13                    & 79.95                    & 77.84                    & 76.08                    \\
                & zeroshot & 78.41                    & 77.34                    & 76.34                    & 77.16                    & 77.34                    & 75.78                    & 74.23                    & 71.51                    \\ \hline
imagenet-r      & map      & 91.04                    & 90.33                    & 91.77                    & 93.66                    & 93.31                    & 91.41                    & 90.72                    & 87.58                    \\
                & zeroshot & 93.79                    & 92.94                    & 94.83                    & 95.81                    & 95.36                    & 94.28                    & 93.54                    & 91.03                    \\ \hline
imagenet-a      & map      & 76.48                    & 65.04                    & 80.19                    & 80.88                    & 84.56                    & 79.77                    & 72.19                    & 64.01                    \\
                & zeroshot & 79.73                    & 69.93                    & 82.28                    & 82.52                    & 86.07                    & 80.61                    & 75.56                    & 65.19                    \\ \hline
imagenet-sketch & map      & 69.08                    & 68.26                    & 68.59                    & 70.93                    & 70.81                    & 67.96                    & 66.79                    & 62.61                    \\
                & zeroshot & 73.44                    & 72.94                    & 71.86                    & 74.79                    & 74.77                    & 71.72                    & 70.73                    & 62.68                    \\ \hline
objectnet       & map      & 66.77                    & 63.4                     & 73.87                    & 71.71                    & 73.28                    & 72.94                    & 70.4                     & 68.18                    \\
                & zeroshot & 71.1                     & 66.01                    & 76.59                    & 73.28                    & 76.83                    & 74.26                    & 73.65                    & 68.8    
                \\ \hline
\end{tabular}
\caption{For ImageNet-1K and ImageNet-V2 datasets, the trained linear classifiers (trained with ImageNet-1K) outperform zeroshot models. However, for datasets  such as ImageNet-A, ImageNet-R, ImageNet-Sketch, and ObjectNet, Zeroshot models consistently outperform trained classifiers, implying that these datasets exhibit larger domain shifts from ImageNet-1K.  }
\label{tab:individual_imagenet}
\end{table}

\section{why BMA’s training-data–determined weights become unreliable under distribution shift, and why OMA’s test-time entropy minimization avoids this failure mode}\label{supp:whywhy}

To isolate the effect of domain shift, we consider one-dimensional linear regression models where all quantities—posterior, marginal likelihood, and posterior predictive distributions—are available in closed form. Let each candidate model be defined by a feature extractor $\phi_l(\dot)$. For any pair $(x,y)$, we have:

\begin{align}
    &\mbox{likelihood}: p(y|\phi_l(x), w_l) = \mathcal{N}(y|\phi_l(x)w_l, \tau^{-1}), \\
    &\mbox{prior}: p(w_l) = \mathcal{N}(w_l|0, \alpha^{-1}), \\
    &\mbox{posterior}: p(w_l|\phi_l(x),y) = \mathcal{N}(w_l| m_l, S_l), \\
\end{align} 
where $m_l=\frac{\tau \phi_l(x) y}{\alpha + \tau \phi^2_l(x)}$ and $S_l=\frac{1}{\alpha + \tau \phi^2_l(x)}$.

Furthermore, 
\begin{align}
    \mbox{log-marginal likelihood}: \log p(y|\phi_l(x)) = \frac{1}{2} \log \alpha + \frac{1}{2} \log \tau - E(m_l) - \frac{1}{2} \log |A_l| - \frac{1}{2} \log (2 \pi),
\end{align} 
where $E(m_l)=\frac{\tau}{2}(y-\phi_l(x)m_l)^2 + \frac{\alpha}{2} m_l^2$ and $A_l = \alpha + \tau \phi^2_l(x)$. 

BMA balances data fit by $\log E(m_l)$, and model complexity by $\log|A_l|$. Crucially, \textbf{this balance is computed purely from the training distribution.}

\paragraph{Where BMA Fails Under Domain Shift}

Consider two candidate models with $\phi_1(x)=x$ and $\phi_2(x)=\frac{1}{x}$. With a broad prior ($\alpha \to 0$), both models can perfectly fit the data making $E(m_1) \to 0$ and $E(m_2) \to 0$, by setting $m_1 = \frac{y}{x}$ and $m_2 = xy$. So, the model preference depends entirely on the complexity term, $\log|A_l|$. 

For training points with $x>1$,
\begin{align}
    A_1 &= x^2 \tau, \\
    A_2 &= \frac{\tau}{x^2}.
\end{align}
At a fixed value of $\tau$, if the training distribution ensures any $x>1$, $\log A_1$ is larger than $\log A_2$, yielding a smaller value of the marginal likelihood for model 1. Consequently, \textbf{model 2 is preferred and will have a higher posterior model weight in BMA.}

Now, consider domain-shifted test inputs $\tilde{x}<1$. We cannot compute $E(m_1)$ and $E(m_2)$ (labels are unknown), and the model preference in terms of the model complexity is now reversed: 
\begin{align}
A_1 &= \tilde{x}^2 \tau, \\  
A_2 &=\frac{\tau}{\tilde{x}^2}.
\end{align}
So under the test distribution, a complexity-based preference would favour \textbf{model 1}. But BMA cannot update this preference because it relies entirely on \textbf{training-data evidence}. This demonstrates the core issue: \textbf{An illustration of how source-evidence weighting and target-entropy weighting can prefer different models.}

This is precisely the situation where BMA fails and test-time adaptation becomes necessary.

\paragraph{Why OMA Succeeds}

Let's turn our attention to OMA. In the case of linear models, each model's posterior predictive distribution is 
$p_l(\tilde{y}|\tilde{x}) = \int p(\tilde{y}|\tilde{x},w_l) \mathcal{N}(w_l| m_l, S_l) d w_l = \mathcal{N}(\tilde{y}| m_l\phi_l(\tilde{x}), \frac{1}{\tau}+\phi^2_l(\tilde{x})S_l)$. 

With model-averaging, the predictive distribution becomes a mixture of Gaussians (MoG) where the mixture weight is denoted by $\beta_l$: 
$p(\tilde{y}|\tilde{x}) = \sum_{l} \beta_l \mathcal{N}(
\tilde{y}| m_l\phi_l(\tilde{x}), \frac{1}{\tau}+\phi^2_l(\tilde{x})S_l)$. 

In OMA, we find $\beta_l$ that minimize the entropy of this prediction. Unfortunately there is no closed form for entropy under the mixture of Gaussians. However, there exists a tight upper bound: 
entropy of MoG $\leq \sum_{l} \beta_l \log \frac{1}{\beta_l} + \sum_{l} \beta_l h(g_l)$, 
where the second term is a weighted sum of each Gaussian's differential entropy, $h(g_l)=\log (\sqrt{2\pi e (\frac{1}{\tau} + \phi^2_l(\tilde{x}) S_l)})= \log (\sqrt{2\pi e}) + \frac{1}{2}\log (\frac{1}{\tau} + \phi^2_l(\tilde{x}) S_l)$. 

In the one-dimensional example above, ignoring the constant term $\log (\sqrt{2\pi e})$, each model's entropy becomes, as $\alpha \to 0$, 
\begin{align}
    h(g_1)&=\frac{1}{2}\log (\frac{1}{\tau}(1 + (\frac{\tilde{x}}{x})^2)), \\
    h(g_2)&=\frac{1}{2}\log(\frac{1}{\tau}(1 + (\frac{x}{\tilde{x}})^2)).
\end{align}

Now if we run the OMA for the domain-shifted test distribution
So, for $\tilde{x}<1$, at a fixed value of $\tau$, $h(g_1)$ < $h(g_2)$, 
so minimizing entropy yield $\beta_1 > \beta_2$. \textbf{Unlike BMA, OMA correctly shifts weight toward model 1—the model with lower test-time predictive variance—because it consults the actual test inputs.}

\textbf{Thus, OMA adapts directly to the test distribution via predictive entropy, whereas BMA is locked to the training distribution and cannot react to a shift.}

\section{Rest of Related Work}\label{supp:rest_related_work}

\paragraph{Predictive ensembling and learned weights.}
Deep ensembles provide a widely studied approach to predictive
uncertainty, including under distribution shift
\citep{deepensembles,ovadia2019can}. Output ensembling is not
restricted to equal weights: BMA uses posterior model probabilities,
while stacking learns combinations through cross-validated
predictive scoring \citep{yao2018stacking}. BMA reduces to uniform
averaging when its model weights are uniform and the component
predictive distributions are unchanged. These weighting principles
are distinct from OMA's target-entropy surrogate. In particular,
source-validation stacking uses labeled source outcomes to learn
weights, whereas OMA adapts its coefficients using unlabeled target
predictions. Avoiding further backbone training is a property of
our adaptation interface, not a distinction from every existing
output-ensemble method.

\paragraph{Black-box adaptation and prediction regularization.}
DINE adapts from single or multiple black-box source predictors by
distilling their outputs into a target model and then fine-tuning it
\citep{liang2022dine}. LAME instead optimizes output assignments
using a Laplacian-adjusted objective and feature-based affinities,
without changing network parameters \citep{boudiaf2022lame}.
OMA neither trains a separate target network nor optimizes an
independent probability vector for every example; it learns one
coefficient per candidate from cached probabilities.
Its prediction-space KL penalty also relates to soft-target matching
in knowledge distillation \citep{hinton2015distilling}. Here, the
reference is a fixed distribution derived from the candidate
predictions, and the adaptable predictor is restricted to their
shared-weight mixture rather than a separately parameterized student.

\paragraph{Entropy-based adaptation and foundation models.}
Tent adapts normalization-related quantities by minimizing predictive
entropy \citep{wang2021tent}, while EATA combines selective entropy
minimization with an anti-forgetting parameter regularizer
\citep{niu2022eata}. SHOT combines information maximization and
pseudo-labeling while adapting a feature extractor with its
classifier fixed \citep{liang2020shot}. In vision-language models,
TPT adapts prompts using augmented test inputs \citep{shu2022tpt},
and TDA performs backpropagation-free adaptation through a dynamic
cache of features and pseudo-labels \citep{karmanov2024tda}.
OMA shares the use of unlabeled target data but restricts adaptation
to mixture coefficients: neither model features nor prompt
parameters are required once probabilities are cached.

\paragraph{Calibration under adaptation.}
Confidence calibration is distinct from predictive accuracy
\citep{guo2017calibration}, and calibration achieved on source data
can deteriorate under distribution shift \citep{ovadia2019can}.
C-TPT explicitly addresses calibration in test-time prompt tuning
through text-feature dispersion \citep{yoon2024ctpt}.
The TTA-VLM benchmark further shows that accuracy gains from
vision-language adaptation can coincide with worse trustworthiness
metrics, motivating evaluation beyond accuracy
\citep{sheng2025ttavlm}. Our study examines this issue for
heterogeneous probability mixtures. The target-reference KL and
reference-weight penalties control adaptation, but do not directly
optimize ECE; we therefore evaluate calibration separately rather
than identifying reduced entropy with calibrated predictions.

\paragraph{Bayesian last layers and parameter-efficient uncertainty.}
Our optional source-informed reference weights connect to Bayesian
last-layer approximations. \citet{pmlr-v119-kristiadi20a} study how
Gaussian weight uncertainty, including last-layer Laplace
approximations, can mitigate overconfidence in ReLU networks.
VBLL develops a deterministic variational formulation for Bayesian
last layers \citep{harrison2024variational}, while Laplace-LoRA
approximates the posterior over low-rank adaptation parameters
\citep{yang2024bayesian}. These approaches model uncertainty within
individual predictors; they do not by themselves learn target-adaptive
weights across heterogeneous models. Their predictive distributions
can nevertheless serve as ensemble components. In our framework,
approximate BMA supplies an optional source-derived reference, while
OMA's target adaptation does not require Bayesian inference or
parameter-posterior integration.

\section{Are the results shown in \tabref{image-ours} statistically significant?}

We divided the training data into three training subsets, where each comes from 33\% of Imagenet-1K training data. We then computed the MAP estimates based on each training subset and computed the prediction accuracy on the validation set, by combining each MAP estimate and zeroshot models. The results are shown in \tabref{stats}. Two things to note: (a) because we are using only 33\% of the original training data, the accuracy of BMA is lower than what we showed in \tabref{image-ours}; and (b) across three training subsets, the accuracy of our method does not change by much, indicating the results we showed in \tabref{image-ours} are not a chance. 

\begin{table}[h]
\begin{center}
\begin{tabular}{  c| c c c c c }
\hline 
& Imagenet-1K & V2 & R & Sketch & A \\
\hline
 training subset 1 & 88.402     & 81.21     &94.74 & 73.478 & 85.743  \\ 
 \hline
training subset 2 & 88.634     & 81.12     &94.847 & 73.308 & 86.39 |\\  
\hline
training subset 3 &  88.542  & 81.36     &94.883 & 73.625 & 85.943 \\
\hline
variance & 0.0090 & 0.0098 & 0.0036 &0.0167 & 0.0734\\
\hline 
\end{tabular}
\end{center}
\caption{Across three training subsets, the accuracy of our method does not change much and the variance across different subsets is relatively small (last row), indicating the results we showed in \tabref{image-ours} are not due to random chance. }\label{tab:stats}
\end{table}
% |  | Imagenet-1K | V2 | R | Sketch | A | 
% | -------- | -------- | -------- |-------- | -------- | -------- |
% | training subset 1  | 88.402     | 81.21     |94.74 | 73.478 |85.743 |
% | training subset 2 | 88.634     | 81.12     |94.847 | 73.308 | 86.39 |
% | training subset 3  | 88.542  | 81.36     |94.883 | 73.625 | 85.943 |

\section{Performance on Other Image Classification datasets (Camelyon17, Flowers 102, and Sun397)}\label{supp:Other_Data}

We use the five foundation models shown in \tabref{open_clip_models_no_imagenet} as feature extractors for relatively simple datasets such as Camelyon17, Flowers102, and Sun397. 

\subsection{Details on the datasets}

\paragraph{Camelyon17 \citep{camelyon}}
Camelyon17-WILDS is part of the WILDS benchmark suite of datasets, containing $455,954$ medical images at $96 \times 96$ resolution. The downstream task is determining whether a given image contains tumour pixels, creating a binary classification task.
The dataset contains $34,904$ test images. 

\paragraph{Flowers102 \citep{Nilsback08}}
Flowers102 contains images of flowers belonging to 102 different categories (i.e., multi-class classification with 102 classes). The training dataset contains at least 40 images for each category, and $1020$ training images. The dataset contains $1020$ test images. Because the test and training image sets are small, the test accuracies between different models do not differ much.  

\paragraph{Sun397 \citep{Xiao2010SUNDL}}
Scene UNderstanding (SUN) database contains 397 categories and 19850 training and 19850 test images (downloaded from \url{https://github.com/open-mmlab/mmpretrain}) for scene recognition tasks.

\subsection{Performance}

Using the five  frozen feature extractors shown in \tabref{open_clip_models_no_imagenet}, we attach a linear classifier and find the MAP estimates for the linear classifier. 

\begin{table}[h]
\centering
\rowcolors{2}{gray!10}{white}
\scalebox{0.7}{\begin{tabular}{@{}l l l r r r c@{}}
\toprule
\textbf{Name in Paper} & \textbf{OpenCLIP Name} & \textbf{Pretraining Data} & \textbf{Params (M)} & \textbf{FLOPs (B)} & \textbf{Memory (G)} & \textbf{Avg. Perf. (38 sets)} \\
\midrule
\textcolor{blue}{fe-1} & ViT-H-14-378-quickgelu         & dfn5b                   & 986.71  & 1054.05 & 4.4   & 0.7079 \\
%%%%%%%%%%%%%%%%%
\textcolor{blue}{fe-2} & ViT-SO400M-14-SigLIP-384       & webli                   & 877.96  & 723.48  & 4.1   & 0.6921 \\
\textcolor{blue}{fe-3} & ViT-bigG-14-CLIPA-336          & datacomp1b              & 2517.76 & 2271.58 & 10.35 & 0.6842 \\
\textcolor{blue}{fe-4} & EVA02-E-14                     & laion2b\_s4b\_b115k     & 4704.59 & 2311.42 & 18.8  & 0.6690 \\
\textcolor{blue}{fe-5} & ViT-H-14-quickgelu             & metaclip\_fullcc        & 986.11  & 381.68  & 4.4   & 0.6671 \\
\bottomrule
\end{tabular}}
\caption{We select $5$ pre-trained open-clip models from \citep{ilharco_gabriel_2021_5143773} as feature extractors for Camelyon17, Flowers102, and Sun397 datasets. }
\label{tab:open_clip_models_no_imagenet}
\end{table}

\begin{table}
\centering
\begin{tabular}{|c|cc|cc|cc|}
\hline
           & Camelyon17 &          & Flowers102 &          & Sun397 &           \\ \hline
Name in Paper & map        & zeroshot & map        & zeroshot & map    & zeroshot \\ \hline 
fe-1          & 91.48      & \textbf{65.71}    & \textbf{99.8}       & 90.59    & \textbf{85.49}  & \textbf{77.09}    \\
fe-2          & 91.73      & 51.65    & 99.61      & \textbf{92.35}    & 85.08  & 75.41    \\
fe-3          & \textbf{92.22}      & 51.81    & \textbf{99.8}       & 87.84    & 85.06  & 76.32    \\
fe-4          & 91.54      & 50.81    & 99.71      & 83.14    & 84.54  & 76.57    \\
fe-5          & 91.49      & 65.53    & 99.61      & 85.59    & 83.43  & 76.53   \\ \hline 
\end{tabular}
\caption{Individual models' performance tested on Camelyon17, Flowers102, and Sun397 datasets. MAP estimates (training with each data's training set) always outperform zeroshot models in these datasets. }
\label{tab:individual_no_imagenet}
\end{table}

In \tabref{individual_no_imagenet}, we show the performance of each model. In all three datasets, trained MAP estimates outperform the zeroshot models. The accuracy of MAP estimates under different feature extractors varies depending on the dataset. For instance, for Camelyon17, fe-3 seems better than others, while for Sun397, fe-1 seems better than others.  
% In the Flowers102 dataset, because the test and training image sets are small, the test accuracies between different models do not differ much. 

\begin{table}[]
\begin{tabular}{cc|c|c|c|}
\cline{3-5}
                                   &            & Cam17          & Flw102         & Sun397                           \\ \cline{1-5} 
\multicolumn{1}{c|}{Zeroshot}      & Output avg & 51.78          & 90.69          & 79.61                            \\ %\cline{2-5} 
\multicolumn{1}{c|}{}              & OMA        &     \textit{70.42}           &             \textit{92.94}    & \textit{79.8}\\ \cline{1-5} 
\multicolumn{1}{c|}{MAP}           & Output avg & 93.63          & 99.80          & 85.82                            \\ %\cline{2-5} 
\multicolumn{1}{c|}{}              & BMA        & \textbf{93.64} & \textbf{99.80} & \textbf{85.84}                   \\ \cline{1-5} 
\multicolumn{1}{c|}{Zeroshot\&MAP} & Output avg & 93.63          & 99.80          & 85.82                            \\ %\cline{2-5} 
\multicolumn{1}{c|}{}              & OMA        & \color{red}\textbf{93.70}      &      \color{red}\textbf{99.83}          & \color{red}\textbf{86.39}              \\ \cline{1-5}
\end{tabular}
\caption{BMA and OMA performances tested on Camelyon17, Flowers102, and Sun397 datasets. Using both MAP and zeroshot models together with our OMA paradigm improves the performance.}
\label{tab:bma_oma_no_imagenet}
\end{table}

In \tabref{bma_oma_no_imagenet}, we show the performance of our algorithms, BMA and OMA, when using zeroshot models and MAP estimates separately and simultaneously.  
In all three datasets, the zero-shot models using our OMA weights improve the performance of output averaging, up to approximately $20\%$ (the column, called Cam17 and the first row).   
In all three datasets, using both MAP and zeroshot models together with our OMA paradigm improves the performance.

\section{Comparison methods for ImageNet-1K and its variant datasets}\label{supp:comparison}

CoCa \citep{yu2022coca}: An encoder-decoder model uses a Vision Transformer (ViT) to encode images and a transformer decoder for text. Unlike standard models, CoCa omits cross-attention in the first half of the decoder for unimodal text, then cross-attends to the image for multimodal representations. This design enables both contrastive and generative objectives. Using both objectives, CoCa is first pre-trained with internet-scale data. These objectives seem the core reason for superior pre-trained models compared to models trained with the contractive objective only (like in our case). For fine-tuning (the value we showed in our Table 3), an additional pooling layer and an extra classification layer were added and fine-tuned for ImageNet-1K while the rest of the model was frozen. Section 4.1 in \citep{yu2022coca}  writes that pretraining CoCa takes about 5 days on 2,048 CloudTPUv4 chips, for 500k steps, roughly corresponding to 5 epochs on JFT. Table 8 in \citep{yu2022coca} shows 200k steps for fine-tuning with frozen features, which seems to indicate 2 days on the same 2,048 CloudTPUv4 chips.

Model Soups \citep{model_soups}: In Model soups, a ViT-G/14 model is first pre-trained with JFT-3B. Then, the model is fine-tuned for ImageNet-1K. Instead of selecting the best model from a hyperparameter sweep during fine-tuning, model soups average the weights of multiple fine-tuned models (See the paper for precise ways to average those weights). Among different ways to ensemble those models, greedy soup (adding models to the pool if they increase the performance) performed the best and we chose to show that result in our Table 3. The authors do not specify how long it takes to fine-tune ViT-G/14 model, while based on the number of parameters of the model we guess the run time is similar to the one mentioned above (CoCa).

EVA02-L \citep{EVA02L}: The EVA02-L model consists of 304M parameters in the vision transformer architecture. What makes the performance of EVA02-L superb is the massive and diverse pre-training data. In the case of results we showed in our Table 3 for EVA02-L, the model was trained with merged 38M data (38 million images from IN-21K, CC12M, CC3M, COCO, ADE20K, Object365 and OpenImages.)

\section{Ablation Study for Image Data}
\label{supp:ablation}

\subsection{Choosing the right prior variance}

In BMA, the prior variance $\alpha$ in \eqref{marginal3} is a hyperparameter we need to tune to find a good MAP estimate.
\tabref{alpha_hyper} shows the ablation study for Camelyon17. For other datasets, we used the same grid search over $\alpha$ and chose the optimal value that maximizes the test accuracy. For Flowers102, it was 10, and for Sun397, it was 100.  
For ImageNet-1K, the optimal prior variance was 80 when using a grid with values of $\{0.01, 0.1, 10, 40, 80, 160\}$. 

\begin{table}[h]
\centering
\begin{tabular}{|c|c|c|c|c|}
\hline
batch size & learning rate & epochs & $\alpha$ & test accuracy \\
\hline 
1000	&0.01	&200	&0.1	&90.15 \\
1000	&0.01	&200	&1.0	&92.21\\
1000	&0.01	&200	&\textbf{10.0}	&\textbf{92.43}\\
1000	&0.01	&200	&50.0	&92.18\\
1000	&0.01	&200	&100.0	&92.15\\
\hline
\end{tabular}
\caption{Hyperparameter search for the prior variance $\alpha$ for Camelyon17}
\label{tab:alpha_hyper}
\end{table}

\subsection{Choosing the right regularization constant}

In OMA, we have a regularization constant $\lambda$ in \eqref{objective} that we need to tune to find a good $\beta$. \tabref{lambda_hyper} shows the optimal $\lambda$ value for ObjectNet when using zeroshot models. Similarly, for other datasets, we use the similar grid search to find their optimal $\lambda$. 
For using both MAP and zeroshot models, for ImageNet-1K, the optimal $\lambda$ was 1.0; for ImageNet-V2, it was 10,000; for ImageNet-R, it was 10; for ImageNet-Sketch it was 1.0; for ImageNet-A, it was 10.0; and for ObjectNet, it was 10,000.

For using zeroshot models, for ImageNet-1K, the optimal $\lambda$ was 1000; for ImageNet-V2, it was 10; for ImageNet-R, it was 1.0; for ImageNet-Sketch it was 1000; for ImageNet-A, it was 10.0; and for ObjectNet, it was 1.

\begin{table}[h]
\centering
\begin{tabular}{|c|c|c|c|c|}
\hline
batch size & learning rate & epochs & $\lambda$ & test accuracy \\
\hline
1200	&0.001	&400	 &0.1	&77.086	
\\
1200	&0.001	&400	 &1.0 &77.121 \\	
1200	&0.001	&400	&10.0 &77.064 \\	
1200	&0.001	&400	&100.0	&78.035 \\	
1200	&0.001	&400	&\textbf{1000.0}	& \textbf{78.069} \\
1200	&0.001	&400	&10000.0	&77.143 \\
\hline
\end{tabular}
\caption{Hyperparameter search for the regularization constant $\lambda$ for ObjectNet}
\label{tab:lambda_hyper}
\end{table}

\subsection{Initial values for $\beta_0$}

In OMA, we have to decide on which initial values we want to use for $\beta_0$ in \eqref{objective}.
When the posterior model weights are available for MAP estimates, we can use these values as $\beta_0$. However, when there is no such initialization available for the case of zero-shot models, we used the log-likelihood of the training data given the model as a proxy to $\beta_0$. 

Recall the model posterior can be computed by identifying $p(\Dat | M_l)$, which we approximated with the point estimate of $\vw_l$ at a zero-shot weight $\vw_l^{zeroshot}$:
\begin{align}
    p(\Dat | M_l) & \approx \int_{\vw_l} p(\Dat |\vw_l, M_l) p(\vw_l|M_l) d\vw_l,\nonumber \\
    &\approx p(\Dat |\vw_l^{zeroshot}, M_l).
\end{align} This is nothing but the first term in \eqref{marginal3}, given a zeroshot model.  

\section{ECE and Acc as a function of regularization strength}

\begin{figure*}[h]
    \centering
    \includegraphics[width=1\textwidth]{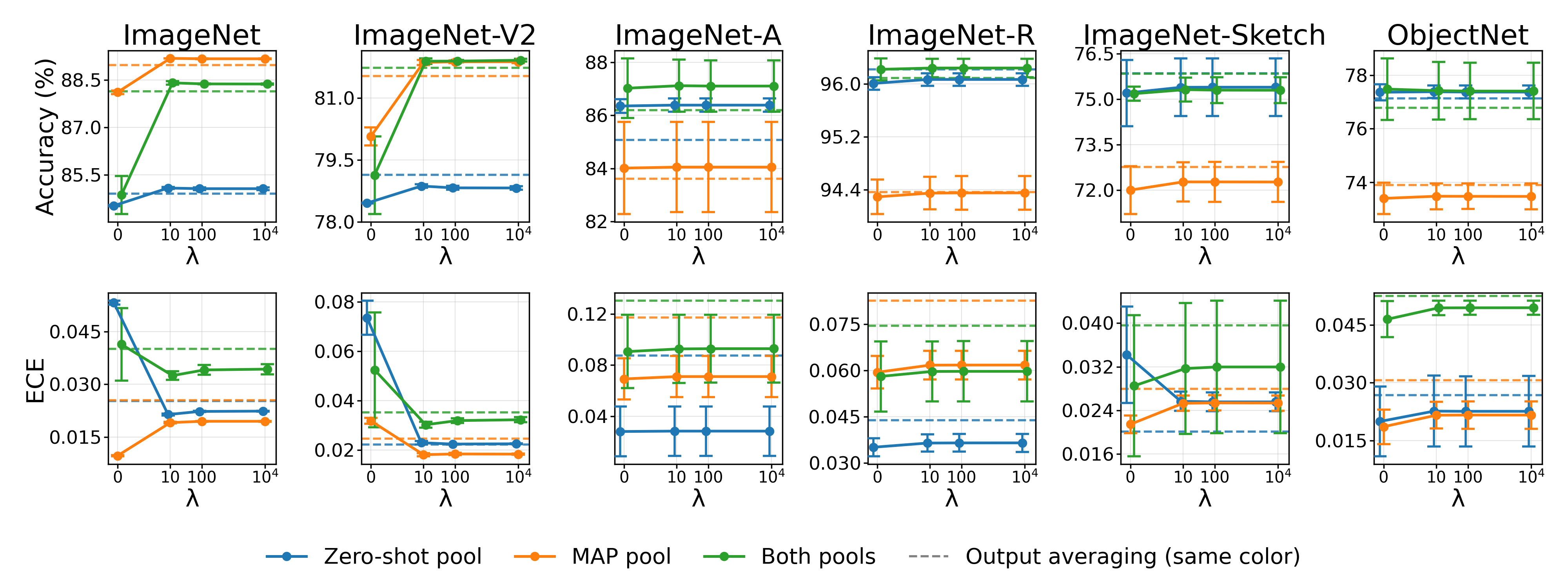}
    \caption{Lambda ablation}
    \label{fig:lambda_ablation_image}
\end{figure*}

\figref{lambda_ablation_image} shows ECE and Accuracy as a function of regularization strength. ImageNet and V2 show similar trends: accuracy increases as $\lambda$ increases, while ECE is generally lowest at $\lambda=10$ (MAP pool for ImageNet is an exception).  A, R, Sketch, and ObjectNet show no particular sweet spot when it comes to accuracy. However, ECE varies with $\lambda$ in Sketch and ObjectNet.  

\section{Required Computation}
\label{supp:required_computation_bma_vs_others}

\textbf{Coca} \citep{yu2022coca}: Pretraining CoCa takes about 5 days (500k steps) on 2,048 CloudTPUv4 chips. Table 8 in \citep{yu2022coca} 
shows fine-tuning with frozen features requires 2 days on the same 2,048 CloudTPUv4
chips.

\textbf{Eva02-L} \citep{EVA02L}: Although not specified, given the massive training data (merged 38M data) and the large model size, the required computation would be more than Coca. Note that the Eva02-L model has a similar architecture to fe-3 in \tabref{open_clip_models}, so one can expect that a fine-tuned fe-3 with ImageNet-1K will produce the accuracy of $90\%$. 

\textbf{Model Soup}: The authors did not specify the runtime. Since this model's performance is closest to ours, for better comparison, we describe the exact runtime of Model Soup on A100 GPUs with 40GB memory.
We considered 20k steps with batch size 512 for fine-tuning ViT-G/14 (over 1 billion parameters) for ImageNet-1K (which is how Model Soups created a single candidate model), which takes 3 days. But this is just for a single model. The Model Soups paper considered 58 models of this size, yielding 174 days of runtime if we have only a single A100 GPU available (which is not unrealistic in academic settings).

\textbf{BMA}: Total runtime for a similarly sized (similar to ViT-G/14) model in BMA is less than 10 hours on a less performing GPU (NVIDIA RTX 4090 GPU with 24GB of memory).
In particular, 
pre-processing (turning images to features for 1,282,167 training points in ImageNet-1K) takes 9 hours; computing the MAP estimate requires less than 2 minutes; and computing the Hessian and the posterior model weight requires roughly 30 minutes for ImageNet-1K. 

\textbf{In summary}, compared to the total runtime for a single model in Model Soup that takes 3 days on a better GPU (A100), the runtime of our method (NVIDIA RTX 4090 GPU) is significantly lower, $86.1\%$ lower than that of Model Soup. If we extrapolate this result for all 58 models in Model Soup versus 8 models in BMA, the total runtime would be 4176 hours for Model Soup and 80 hours for BMA, yielding a $98.0843\%$ decrease in the total runtime. However, the accuracy of each model is comparable: Model Soup on ImageNet-1K achieves $90.94$, and our BMA achieves $89.24$.

\section{Datasets used for Text Classification}\label{supp:text_classification}

Following \citep{pmlr-v162-wortsman22a}, we use the following four text classification datasets from the GLUE benchmark \citep{wang2018glue}.

\subsection{Microsoft Research Paraphrase Corpus} This dataset, often called \textit{MRPC},   
contains pairs of sentences, labelled as either nearly semantically equivalent, or not \citep{dolan2005automatically}. We can view this as a binary classification dataset. The training set consists of 3700 and the validation set of 409 samples. Typically models are evaluated using the average of $F_1$ and accuracy.
    
\subsection{Recognizing Textual Entailment} This dataset, often called \textit{RTE} contains pairs of sentences \citep{wang2018glue}, from a series of datasets \citep{dagan2005pascal, bar2006second, giampiccolo2007third, bentivogli2009fifth}. The task is to predict whether the first sentence (the premise) entails or contradicts the second sentence (the hypothesis).  The training set consists of 2500  samples and the validation set of 277 samples. Since we can view this as a binary classification dataset, models are often evaluated in terms of classification accuracy. 
    
\subsection{Corpus of Linguistic Acceptability}
This dataset, often called \textit{CoLA}, contains sentences labelled as either grammatical or ungrammatical \citep{warstadt2018neural}. Models are often evaluated on Matthews correlation (MCC)  \citep{matthews1975comparison} (ranges between $-1$ and $1$). The training set consists of 8600 samples and the validation set consists of 1043 samples.
     
\subsection{Stanford Sentiment Treebank} 
This dataset, often called \textit{SST-2}, contains sentences labelled as expressing \textit{positive} or \textit{negative} sentiment, collected from movie reviews \citep{socher2013recursive}. Models are often evaluated in terms of classification accuracy. The training set consists of 67,000 samples and the validation set consists of 873 samples.

\section{Checkpoints we used for Text Classification Experiments}
\label{supp:checkpoints}

To access checkpoints, we used the following words to search for these four models 
\textit{'google-bert/bert-base-cased', 'google-bert/bert-large-cased', 'funnel-transformer/xlarge-base' 'funnel-transformer/xlarge'} in \url{https://huggingface.co/models}.

\section{Implementation}
While we plan to publish our code upon the publication of our manuscript, we submit our code as a zipped supplementary file. 

Our code has two folders: Image Classification and Text Classification. 

In Image Classification, 
\begin{itemize}
    \item \textit{preprocess.py}: This script can transform images into feature representations using selected foundation models as feature extractors.
    \item \textit{reorganized\_imagenet\_feats.py}: This script reorganizes image features so that they match corresponding labels. Intended to be used for ImageNet-1K. 
    \item \textit{reorganized\_imagenet\_OOD\_datasets\_feats.py}: The same as above, but for ImageNet OOD datasets.
    \item \textit{training\_with\_preprocessed\_data.py}: This script attaches a linear head to frozen foundation models and finds the MAP estimate of the linear head. 
    \item \textit{hessian\_computation\_with\_p\_1minus\_p.py}: This script computes the block diagonal Hessian and returns the model posterior weights for ImageNet-1K data.
    \item \textit{predict\_given\_zeroshot\_individually\_trained\_weights.py}: This script computes the prediction performance using zeroshot and MAP estimates together or separately using output averaging or the model posterior weights.
    \item \textit{test\_optimization.py}: This scripts perform the OMA experiments. 
\end{itemize} The rest of the files in the Image Classification folder are intended for Camelyon17, Flowers102, and Sun397 datasets.

In Text Classification,  the base code we used is from \url{https://github.com/huggingface/transformers/tree/main/examples/pytorch/text-classification}, which we did not include in the folder.  Using the base code, we fine-tuned the four models mentioned in \suppsecref{checkpoints} first. We then ran our script \textit{test\_optimization.py} to perform the OMA experiments.

\section{Ablation for Text Classification}
\label{supp:ablation_text}

\begin{figure}[H]
    \centering
    \includegraphics[width=\linewidth]{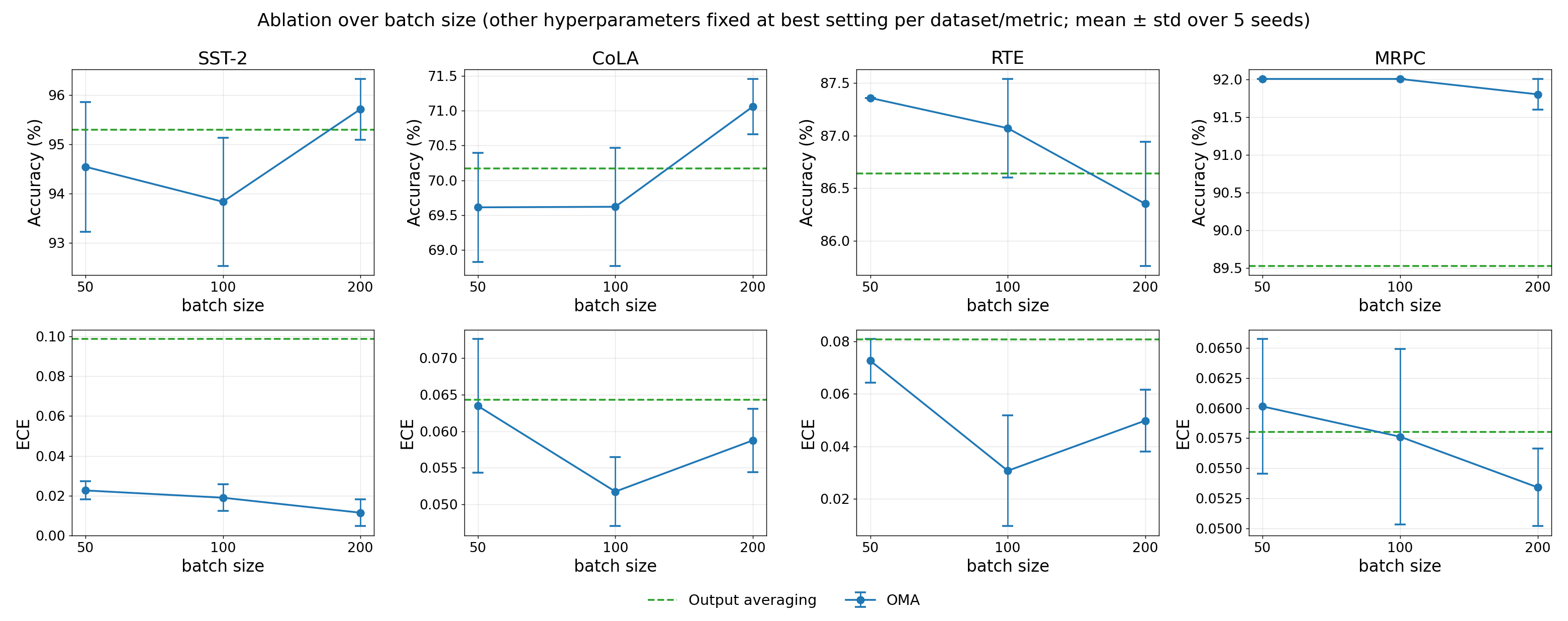}
    \caption{Ablation over batch size.}
    \label{fig:abl_bs}
\end{figure}

\begin{figure}[H]
    \centering
    \includegraphics[width=\linewidth]{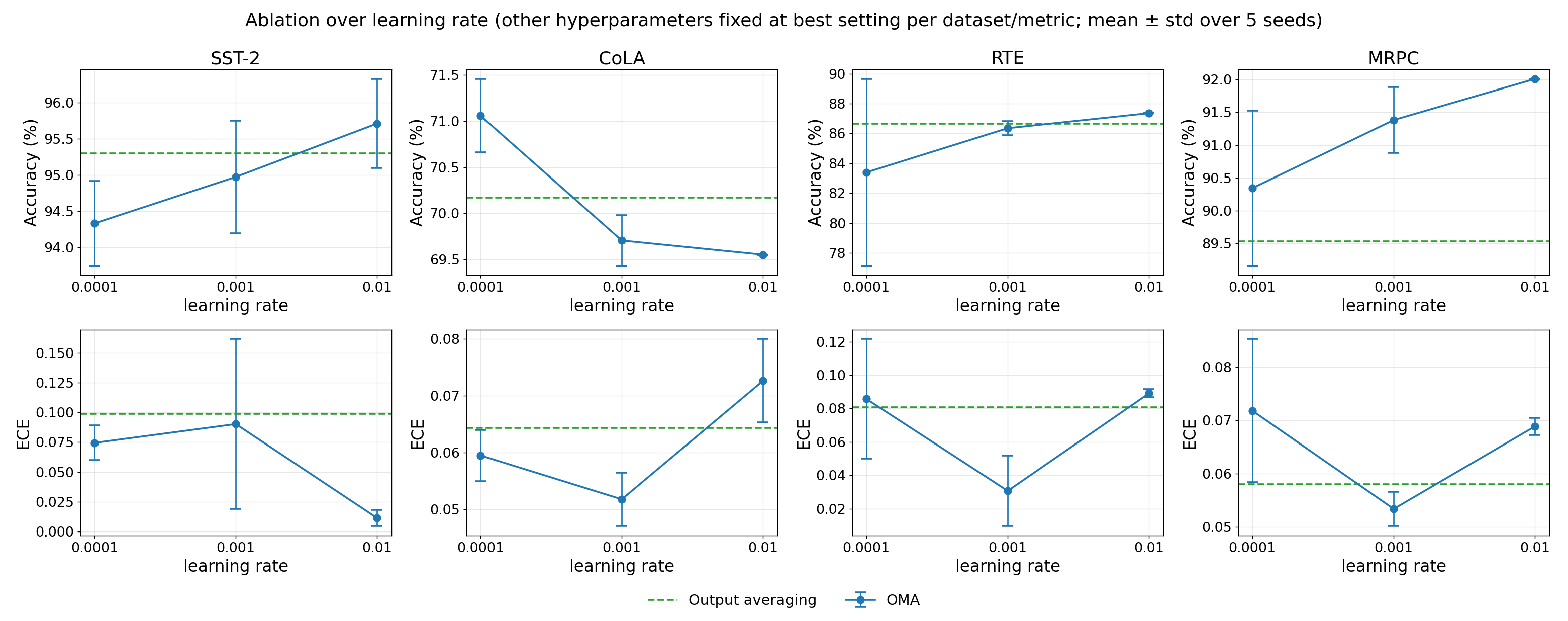}
    \caption{Ablation over learning rate.}
    \label{fig:abl_lr}
\end{figure}

\begin{figure}[H]
    \centering
    \includegraphics[width=\linewidth]{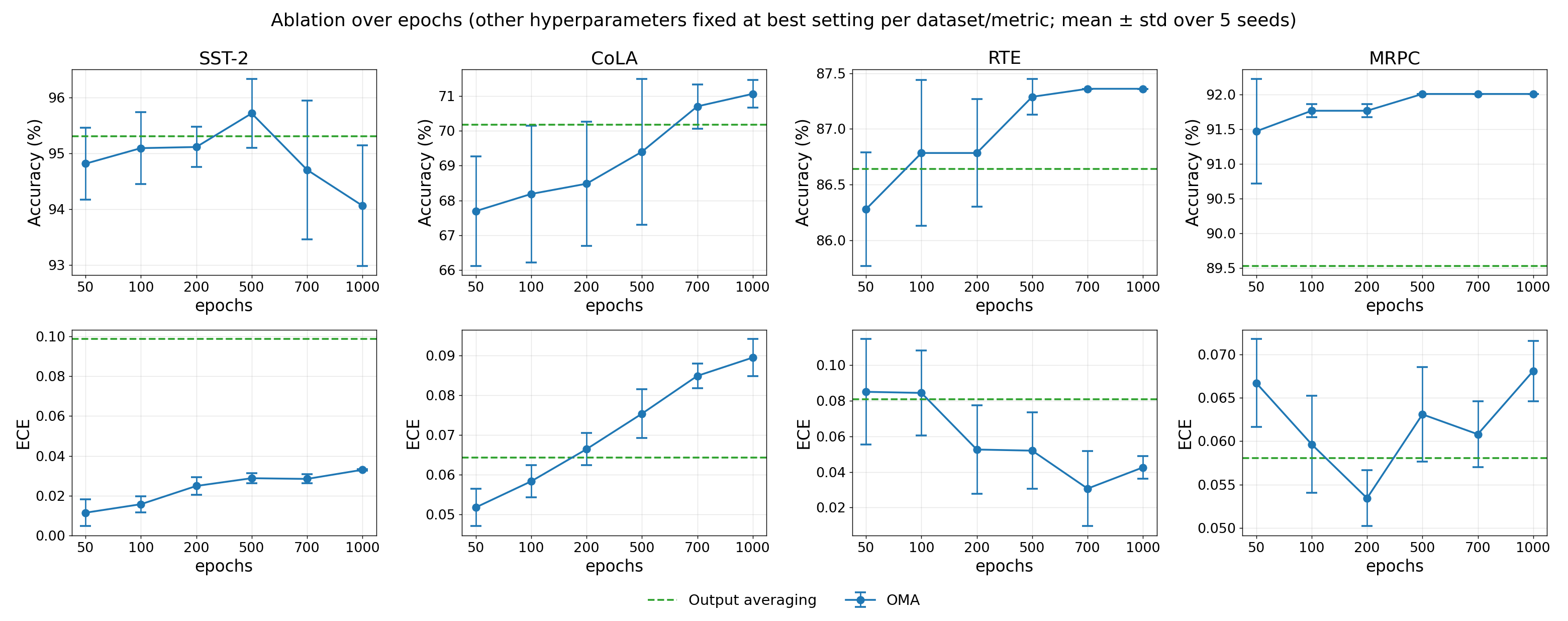}
    \caption{Ablation over the number of epochs.}
    \label{fig:abl_eps}
\end{figure}

\begin{figure}[H]
    \centering
    \includegraphics[width=\linewidth]{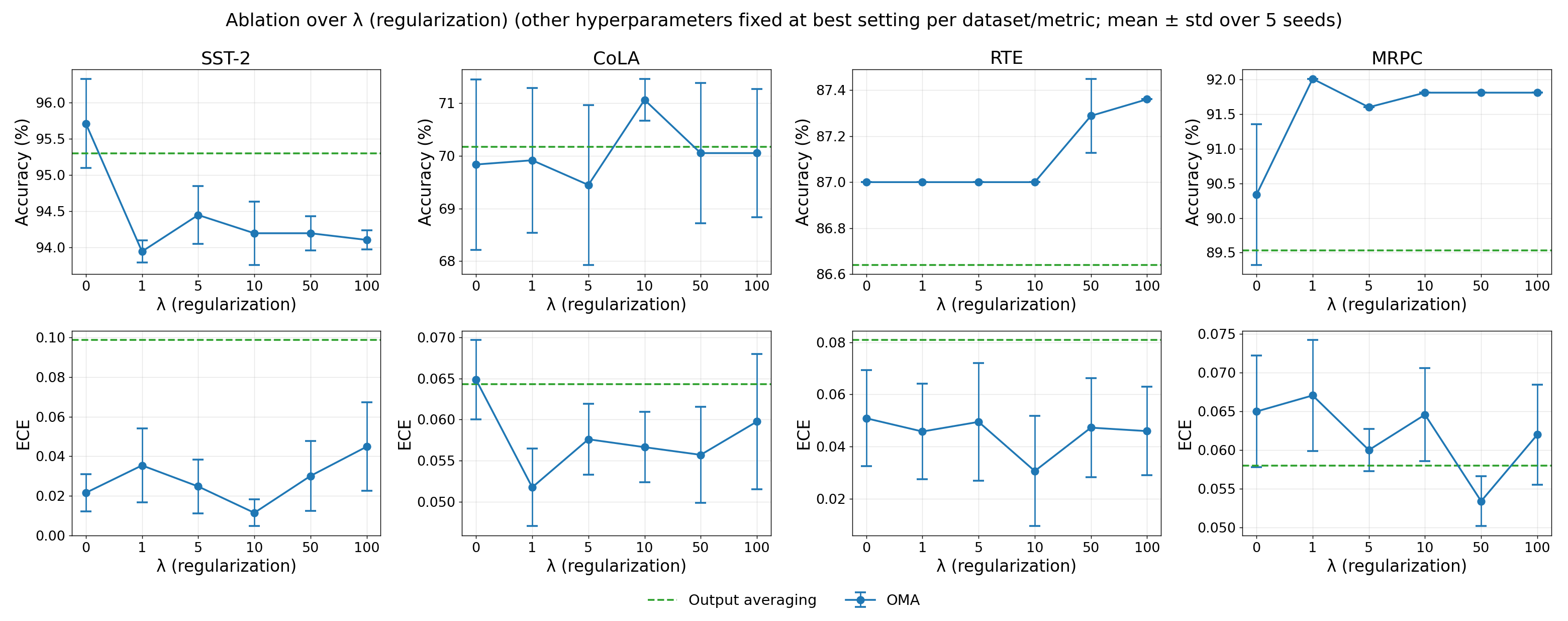}
    \caption{Ablation over the regularization constant $\lambda$.}
    \label{fig:abl_lam}
\end{figure}

Figures~\ref{fig:abl_bs}--\ref{fig:abl_lam} show the effect of each hyperparameter on the accuracy (top row) and ECE (bottom row) of OMA on SST-2, CoLA, RTE, and MRPC. Each point is OMA averaged over 5 seeds at a fixed set of other hyperparameters, and error bars show the standard deviation over 5 seeds. The dashed green line is output averaging, with the shaded band showing its standard deviation over seeds.

\section{How to do text classification without an extra classifier}\label{supp:bert_classification}
In BERT, for instance, the tasks are cast as a masked language modelling problem. For linguistic acceptability, an example is to predict “A) acceptable” for the inputs “sentence: model soups are grammatical. | options: A) acceptable B) unacceptable | label: [MASK] [MASK] [MASK]”. For evaluation, we select which of the options is given the highest probability according to the model.

\section{Comparison of runtime and accuracy of sampling-based approaches vs the proposed Hessian-based approach}\label{supp:comparison_MC}
% Q2 Did you get a chance to run some low-dimensional experiments to give a comparison of runtime and accuracy of sampling-based approaches vs the proposed Hessian-based approach?:

% A2 This is a very good suggestion. 
We downloded the Iris dataset \citep{https://doi.org/10.1111/1740-9713.01589}, which contains four features and three classes. If we use a linear classifier, the weight matrix is simply four by three, making 12-dimensional parameter space. Here, we train the linear classifier and compute the log-marginal-likelihood using the Monte Carlo Integration (to compute \eqref{marginal}) or using the Laplace Approximation (to compute \eqref{un_norm_log_post}).

\tabref{acc_large_var} shows the runtime and the approximated value for \eqref{marginal} using Monte Carlo (MC) integration when the variance is relatively large. Note that Laplace Approximation (LA) gives -75.89 and took 0.045 sec on the same device (We used a free version of Colab for this experiment). As the sample size increases, because the prior variance is relatively large, the values by the MC integration slowly approach the value by LA. However, the runtime increases exponentially as the number of samples increases.

\begin{table}
    \centering
    \begin{tabular}{c|c|c}
        number of samples&	runtime (seconds)	&log-marginal-likelihood \\ \hline
200,000	&26.4&	-79.68\\
400,000	&54.97&	-79.02\\
800,000	&106.62&	-78.31\\
1,600,000&	214.26&	-77.55\\
3,200,000&	436.79&	-76.88\\
6,400,000&	862.89&	-76.18\\
12,8000,000&	1712.4&	-75.51\\
    \end{tabular}
    \caption{When variance is large}
    \label{tab:acc_large_var}
\end{table}

On the other hand, \tabref{acc_small_var} shows the accuracy when the variance is relatively small, where the MC integration converges fast. The log-marginal-likelihood using LA is -92.51.

\begin{table}
    \centering
    \begin{tabular}{c|c|c}
number of samples&	runtime&	log-marginal-likelihood\\ \hline
1,000&	0.13&	-93.57\\
10,000&	1.23&	-94.86\\
100,000&	13.23&	-94.82\\
1,000,000	&128.62&	-94.83\\
10,000,000&	1181&	-94.82
    \end{tabular}
    \caption{When variance is small}
    \label{tab:acc_small_var}
\end{table}
% [2] A. Unwin, K. Kleinman. The Iris data set: In search of the source of virginica, 2021

\section{Possible ways to reduce the inference cost}\label{supp:remedy}

Quantization replaces high-precision floating-point numbers (like 32-bit or 16-bit) with lower-precision ones (e.g., 8-bit integers or even 4-bit integers), albeit at the cost of slightly reduced accuracy. There are multiple types of quantization, such as Post-Training Quantization (PTQ) (quantizing the model after it has been trained, without requiring further training or fine-tuning) \citep{hong2025stablequantlayeradaptiveposttraining}; Dynamic Quantization (only the weights are quantized before inference, and activations are quantized on the fly during inference); and Static Quantization (both weights and activations are quantized before inference, often requiring a calibration step with a sample of the inference data) \citep{10.5555/3618408.3619993}. Using some of these quantization methods, we can reduce the model size, making it easier to store and deploy, enabling faster inference.

Model pruning (post-training) is also a popular choice to reduce inference time costs and memory requirements to store large models \citep{sun2024a}.

Applying these techniques can undoubtedly improve the inference cost in any model output merging techniques (not just our method). 
% We did not pay any particular attention to these approaches, as we see this as somewhat orthogonal to our main contribution.
% However, we can certainly add these methods to our final version to show improved inference time costs.

% [1] Hong et al, StableQuant: Layer Adaptive Post-Training Quantization for Speech Foundation Models, ICASSP 2025

% [2] Xiao et al, SmoothQuant: Accurate and Efficient Post-Training Quantization for Large Language Models, ICML 2023

% [3] Sun et al, A SIMPLE AND EFFECTIVE PRUNING APPROACH FOR LARGE LANGUAGE MODELS, ICLR 2024

\section{Details on LoRA and Bayesian LoRA training}\label{supp:details_lora}

We used 20 epochs for the MRPC dataset, and 10 epochs for the rest of the dataset. We set the rank to 8 and the alpha value to 32 throughout the whole experiment. We set the batch size to 32 and the learning rate to 2e-5.

% Note: You can choose whether to include you appendices as part of the main submission file (here) OR submit them separately as part of the supplementary material. The preferred option is to include appendices as part of the main submission file. It is the authors' responsibility that any supplementary material does not conflict in content with the main paper (e.g., the separately uploaded additional material is not an updated version of the one appended to the manuscript).

% \section{Formatting Instructions}

% The appendices follow the same formatting instructions as in the main paper.
% The only difference is that the supplementary material must be in a \emph{single-column} format.

% Note that reviewers are under no obligation to examine your supplementary material.

% \section{Missing Proofs}

% The supplementary materials may contain detailed proofs of the results that are missing in the main paper.

% \subsection{Proof of Lemma 3}

% \textit{In this section, we present the detailed proof of Lemma 3 and then [ ... ]}

% \section{Additional Experiments}

% If you have additional experimental results, you may include them in the supplementary materials.

% \subsection{Effect of the Regularization Parameter}

% \textit{Our algorithm depends on the regularization parameter $\lambda$. Figure 1 below illustrates the effect of this parameter on the performance of our algorithm. As we can see, [ ... ]}

\end{document}